\documentclass[10pt,twocolumn,letterpaper]{article}
\pdfoutput=1
\usepackage{iccv}
\usepackage{booktabs}
\usepackage{times}
\usepackage{epsfig}
\usepackage{float}
\usepackage{graphicx}
\usepackage{amsmath}
\usepackage{amssymb}
\usepackage{subfig}
\usepackage[dvipsnames]{xcolor}
\DeclareMathOperator*{\argmin}{arg\,min} 

\usepackage[pagebackref=true,breaklinks=true,colorlinks,bookmarks=false]{hyperref}
\usepackage[capitalize]{cleveref}

\iccvfinalcopy       

\def\iccvPaperID{9545} 

\ificcvfinal\fi

\begin{document}

\title{Toward a Visual Concept Vocabulary for GAN Latent Space}

\author{Sarah Schwettmann\textsuperscript{1,2},  Evan Hernandez\textsuperscript{2},  David Bau\textsuperscript{2},  Samuel Klein\textsuperscript{3},  Jacob Andreas\textsuperscript{2},  Antonio Torralba\textsuperscript{2}\\
\textsuperscript{1}MIT BCS, \textsuperscript{2}MIT CSAIL, \textsuperscript{3}MIT KFG\\
{\tt\small \{schwett, dez, davidbau, sjklein, jda, torralba\}@mit.edu}
}

\maketitle
\ificcvfinal\thispagestyle{empty}\fi  

\begin{abstract}
A large body of recent work has identified transformations in the latent spaces of generative adversarial networks (GANs) that consistently and interpretably transform generated images. But existing techniques for identifying these transformations rely on either a fixed vocabulary of pre-specified visual concepts, or on unsupervised disentanglement techniques whose alignment with human judgments about perceptual salience is unknown. This paper introduces a new method for building open-ended vocabularies of primitive visual concepts represented in a GAN's latent space. Our approach is built from three components: (1) automatic identification of perceptually salient directions based on their \emph{layer selectivity}; (2) human annotation of these directions with free-form, compositional natural language descriptions; and (3) decomposition of these annotations into a visual concept vocabulary, consisting of distilled directions labeled with single words. Experiments show that concepts learned with our approach are reliable and composable---generalizing across classes, contexts, and observers, and enabling fine-grained manipulation of image style and content.

\end{abstract}

\section{Introduction} 

GANs~\cite{goodfellow2014generative} map latent vectors $\mathbf{z}$ to images $\mathbf{x}$. Past work has found that directions in this latent space can encode specific aspects of image semantics: StyleGAN trained on bedrooms, for example, contains a direction such that moving most $\mathbf{z}$ in that direction causes indoor lighting to appear in the associated image~\cite{yang2020semantic}. 
However, current methods for identifying these directions are \emph{ad hoc}, capturing only a limited set of human-salient dimensions of variation.
In this paper, we describe how to construct
more expressive and diverse sets of meaningful image transformations—a visual concept vocabulary—by decomposing freeform language descriptions of GAN transformations.

\begin{figure}[t!]
\begin{center}
   \includegraphics[width=\linewidth]{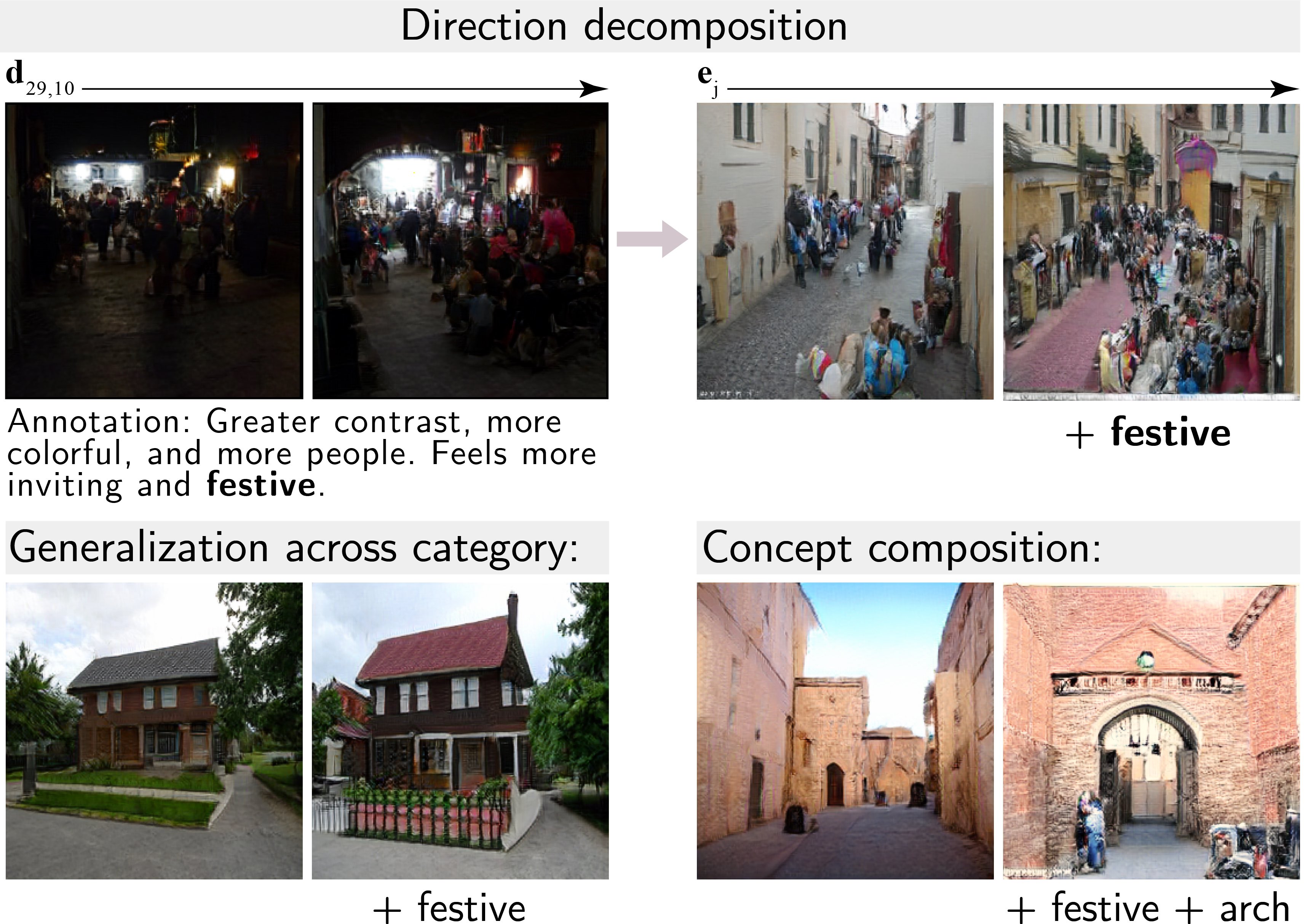}
\end{center}
   \vspace{-10px}
   \caption{Building a visual concept vocabulary. First, we generate directions that preserve most of the structure and content in the image. Then we use human annotations to decompose them into directions that correspond to a single salient concept. Finally, we show the decomposed directions generalize across starting representations and input classes, and can be composed to construct compound directions.}
\label{fig:festive}
\end{figure}

Consider trying to find a direction that makes an outdoor market more \emph{festive} (Figure~\ref{fig:festive}). The GAN latent space is too large to make random search feasible, while supervised approaches cannot 
verify if the desired direction is present ~\cite{jahanian2020steerability, goetschalckx2019ganalyze, yang2020semantic, shen2020interpreting}. Unsupservised approaches ~\cite{harkonen2020ganspace, peebles2020hessian, shen2020closedform, voynov2020unsupervised} may not discover a \emph{festive} direction, since the model's principal components do not necessarily capture changes that are most visually salient to humans. 

To improve our understanding of the kinds of interpretable semantic transformations encoded in GAN latent space, we propose a new approach for building an open-ended glossary of primitive, perceptually salient directions from the bottom up. Our approach is built from three components:

\begin{enumerate}
    \item A new procedure for generating perceptually salient directions based on \textbf{layer selectivity}. The resulting directions make meaningful local changes to a scene but are still non-atomic.
    \item A data collection paradigm in which human annotators directly \textbf{label directions} with their semantics, which are complex and compose multiple concepts to describe visual transformations. 
    \item A new \textbf{bag-of-directions model} which automatically decomposes these annotations into a glossary of ``primitive'' visual transformations associated with single words.
\end{enumerate}

Because our method covers the breadth of the GAN latent space, it enables reliable image editing with a relatively open-ended vocabulary. 
We also show how our vocabulary supports generalization to novel compositions and transfer across classes. Code, data, and additional information are available at \href{visualvocab.csail.mit.edu} {visualvocab.csail.mit.edu}.

\section{Related work}

Our approach is inspired by recent success in discovering latent vectors that capture individual dimensions of semantic variation in images \cite{jahanian2020steerability, harkonen2020ganspace, goetschalckx2019ganalyze, yang2020semantic}. To the best of our knowledge, ours is the first attempt to systematically catalog the set of human-interpretable concepts represented inside a generator's latent space.

\paragraph{Interpreting GANs.} GANs excel at capturing the rich visual structure of images---raising the question of what internal representations they leverage to do so, and the extent to which these representations overlap with dimensions of variation that humans recognize and find meaningful in visual scenes. Early work \cite{radford2015unsupervised} on GANs discovered latent vectors that encode semantically meaningful representations at different levels of abstraction. A subsequent approach \cite{bau2018gan} to the interpretation problem focused on indiviual units, and used a pretrained segmentation network \cite{xiao2018unified} to identify sets of units in intermediate layers whose feature maps closely match the semantic segmentation of a particular object class. Related work identified concepts \emph{not} learned by GANs \cite{bau2019seeing} by comparing the distribution of segmented objects in generated images with the target distribution in the training set.  These  approaches are constrained in the sets of concepts they could possibly identify, which  are limited to the object classes represented inside the segmentation model. In addition to objects, GANs have also been shown to contain internal representations that determine spatial layout \cite{park2019semantic, zhu2018generative, Bau_2019}, and other higher-order scene attributes, inlcuding memorability and emotional valence \cite{goetschalckx2019ganalyze}. While these approaches have made it possible to control specific aspects of image output, looking for a predetermined set of concepts limits what can be learned about what a GAN is able to represent. Our approach to interpretation aims to be more data-driven: by building \emph{shared vocabularies}, represented by GANs and salient to humans, from the ground up.

\paragraph{Supervised direction search.} If concepts to search for are known, and attribute annotations are available, vector directions in latent space can be discovered using supervised classifiers \cite{karras2019style,shen2020interpreting}. When attribute annotations are not available, image classifiers can be used \cite{yang2020semantic}, or a separate model can be trained \cite{jahanian2020steerability,plumerault2020controlling, denton2020image}. However the former is limited to concepts captured by the classifier, and the latter is limited to simple predetermined visual concepts, such as camera angle. Our method does not assume the concepts to search for are known ahead of time.

\paragraph{Unsupervised direction search.} 
Other recent approaches use unsupervised methods for discovering interpretable dimensions in GAN latent space and feature space  \cite{harkonen2020ganspace,voynov2020unsupervised, peebles2020hessian, wu2020stylespace}. These methods make use of the known disentanglement of many GAN representations \cite{bau2018gan}. One such method---GANSpace---discovers latent directions for image manipulation by identifying principal components of feature tensors on the early layers of GANs, and transferring the basis to latent space by linear regression \cite{harkonen2020ganspace}. However, the visual content of most of these transformations is unknown, as only a handful of examples have been labeled by the authors after the fact. Furthermore, this direction-generation procedure is limited to finding disentangled principal components of the model's representation, while many other directions salient to humans may lie outside this set. 

Where related work applies ad-hoc labels to directions discovered with such unsupervised methods, we introduce a bottom-up method for discovering directions associated with concepts, in the case when the set of concepts to search for is not known \textit{a priori}. A primary contribution of our method is that it does not require visual concepts to be perfectly disentangled \emph{before} labeling.

\section{Projecting visual concepts into latent space}
\label{sec:Method}

\begin{figure*}
\begin{center}
\includegraphics[width=.9\linewidth]{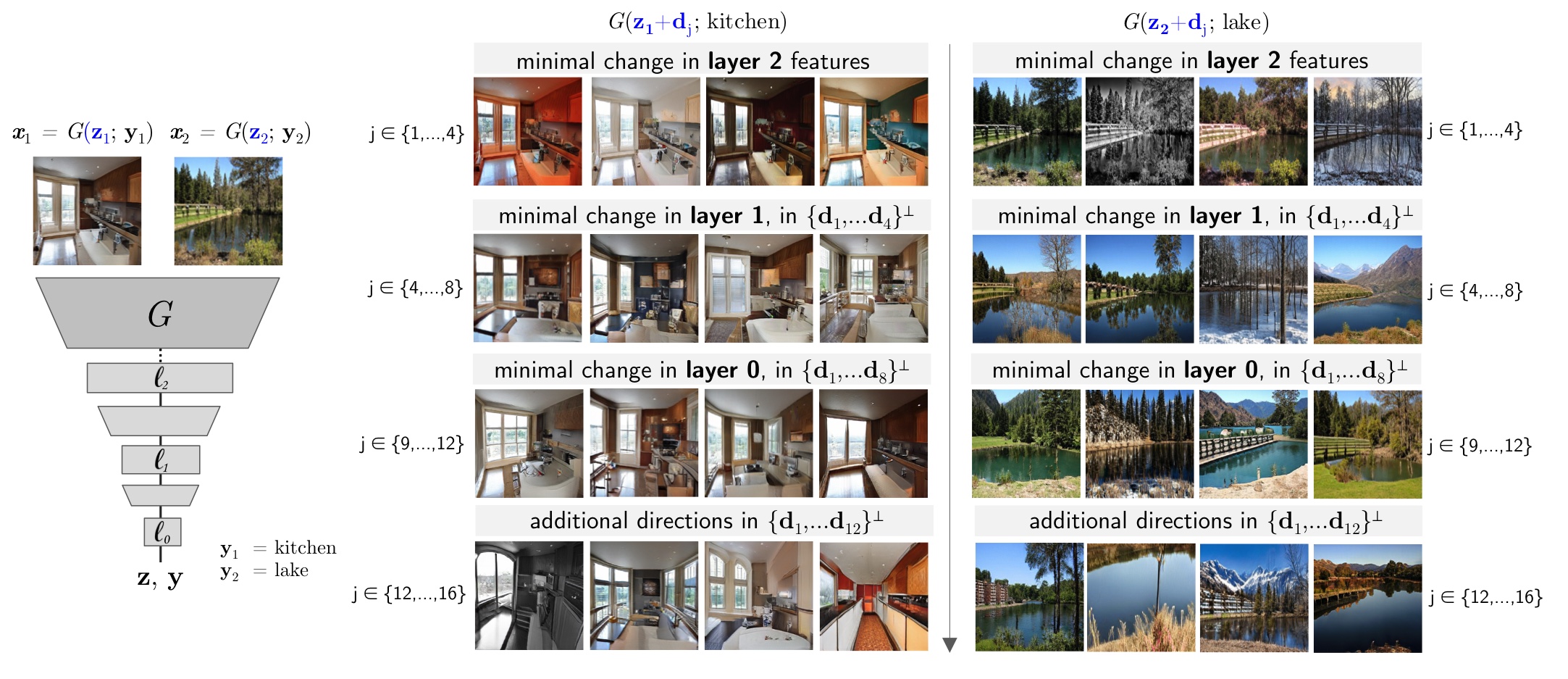}
\end{center}
   \vspace{-20px}
   \caption{Examples of layer-selective directions. Directions are generated by minimizing the change in a layer with respect to the direction, subject to a norm constraint. Our procedure selects a set of ${n}$ LSDs for each layer, one layer at a time, orthogonal to those already selected.}
\label{fig:gendir}
\end{figure*}

Our goal is to distill dimensions of variation in GAN latent space $\mathcal{Z}$ that capture primitive visual transformations in image space. We begin by generating a set of test directions in multiple image classes, where transformation along those directions is constrained to be minimal in a subset of the layers' feature representations, but potentially semantically complex (\cref{sec:directiongeneration}). Next, we synthesize sequences of images transformed along each test direction in $\mathcal{Z}$ latent space, and ask human annotators to describe the corresponding visual changes (\cref{sec:annotation}). Because these directions are generated without preselecting for particular concepts, they act as a screen upon which viewers project the gradients of perceptual change they find most salient. We use the prevalence of repeated terms and their association with different transformations to infer a set of visual concepts represented in the latent space, and associated directions that change the perceived presence of each concept (\cref{sec:Distilled}). 

Experiments in the remainder of this paper use the BigGAN architecture~\cite{brock2019large}, a class-conditional model pretrained on the Places dataset~\cite{bolei2017places}, which includes visual scenes from 365 unique classes. However, our approach is relatively model-agnostic. We show generalization to BigGAN trained on the ImageNet dataset \cite{imagenet_cvpr09} in the supplement.

\subsection{Selecting directions for annotation}
\label{sec:directiongeneration}
A generator $G$ maps latent code $\mathbf{z}$ and class vector $\mathbf{y}$ into image space, synthesizing $\mathbf{x} = G(\mathbf{z};\mathbf{y})$. The image $\mathbf{x}$ can be manipulated along a visual dimension by transforming the vector $\mathbf{z}$ along the corresponding direction $\mathbf{d}$ in the latent space: $\mathbf{x^*} = G(\mathbf{z} + \mathbf{d};\mathbf{y})$. This correspondence between directions in visual and latent space lies at the heart of the problem we wish to solve. For a given model, we want to learn embeddings in the latent space $\mathcal{Z}$ of transformations that are salient to human observers in \emph{visual space}. However, we cannot begin by defining an objective where $\mathbf{d}$ is optimized to produce a discriminable transformation in $\mathbf{x}$, such as in ~\cite{jahanian2020steerability, goetschalckx2019ganalyze,schwettmann2020latent,yang2020semantic}, because we wish to avoid pre-committing to a fixed vocabulary of visual concepts.

\paragraph{Layer-selective directions (LSDs).} To generate directions for annotation, we sample the space of salient perceptual transformations for different $\mathbf{z}$. Our goal is to collect a direction annotation dataset that is both \textit{diverse} and \textit{specific}---capturing a broad set of concepts, where the same concept is reliably associated with a particular direction across images and observers. Thus for a given $\mathbf{z}$, we seek directions that make minimal, meaningful perceptual changes at different levels of abstraction.

Randomly sampled directions tend to alter many visual features, at many levels of resolution, all at once. To constrain a direction $\mathbf{d}$ (of fixed magnitude) to make a smaller number of specific, recognizable changes in the image output $G(\mathbf{z}+\mathbf{d};\mathbf{y})$, we can search for a $\mathbf{d}$ that minimizes change in the feature representation of an intermediate layer of $G$. Denote by $G_{\ell}$ the first $\ell$ layers of $G$.  Then the feature map for that layer is a computed as follows:
\begin{equation}  
g_\ell = G_{\ell}(\mathbf{z},\mathbf{y})
\end{equation}
Let $g_\ell^{*}$ be the output of layer $\ell$ when we add $\mathbf{d}$ to $\mathbf{z}$:
\begin{equation} \label{eq1}
g_\ell^{*} = G_{\ell}(\mathbf{z} + \mathbf{d},\mathbf{y})
\end{equation}

We constrain change in a layer's representation by defining a layer regularizer that minimizes $\left \| g_{\ell}^* - g_{\ell} \right \|^2$ for some layer $\ell$. To generate a diverse set of directions $\mathbf{d}_{j,\ell}$ that meet this objective for layer $\ell$, we sample random vectors $\mathbf{d}$ and then apply gradient descent to each sample to optimize the latent direction $\mathbf{d}_{j,\ell}$ to minimize change in $g_\ell^{*}$, where $\mathbf{d}_{j,\ell}$ is constrained to have unit norm.  We call a direction optimized in this way a \emph{layer-selective direction}.

Different layers of $G$ control features in the image output at different levels of resolution, with later layers controlling more fine-grained features ~\cite{bau2018gan,yang2020semantic}. Therefore, to construct a set of LSDs that encompasses diverse image transformations, when sampling vectors $\mathbf{d}_{j,\ell}$ at layer $\ell$ we add the further constraint that the samples be orthogonal to LSDs for other layers. Formally, our objective becomes:
\begin{align} \label{eq2}
\mathbf{d}_{j,\ell}  = & \argmin_{\mathbf{d} \in U_{\ell}} \left \| g_\ell^{*} - g_\ell   \right \|^2 \\
\text{where }\; U_{\ell}  = & \{ \mathbf{d} \; \text{such that} \; ||\mathbf{d}|| = 1 \text{ and } \nonumber \\
&\quad  \mathbf{d} \perp \mathbf{d}_{j',\ell'} \text{ for all } j' \text{ and } \ell' > \ell\}
\end{align}

We begin by sampling $n$ LSDs at the last layer $\ell$ and then proceed to find orthogonal directions selective for earlier layers. This procedure is analogous to Gram-Schmidt orthogonalization and picks directions that lie along mutually orthogonal subspaces of $\mathcal{Z}$, with transformations in each subspace corresponding to image changes at different levels of abstraction. Finally, we generate a set of $n$ additional directions that are orthogonal to all LSDs, to capture types of image transformation that were excluded by the layer-selective process. Examples of directions generated using this method are visualized in Figure~\ref{fig:gendir}.


\subsection{Collecting direction annotations}
\label{sec:annotation}
We apply the method described in \cref{sec:directiongeneration} to 64 randomly selected $\mathbf{z}$ to generate 20 layer-selective directions $\mathbf{d}_{j}$ per $\mathbf{z}$, for a total of 1280 directions. 
For each $\mathbf{z}_i$, the image $G(\mathbf{z}_i)$ is transformed along each direction $j$ by passing a modified $\mathbf{z}_i$ through the generator: $G(\mathbf{z}+\alpha \mathbf{d}_{i,j})$. 
The transformation is visualized in an image pair: $[G(\mathbf{z}_i), G(\mathbf{z}_i+\alpha \mathbf{d}_{i,j})]$, where $\mathbf{d}$ has unit norm. 
To create images for annotation, we set the scaling term $\alpha = 6$. 

\begin{figure}[H]
\begin{center}
\includegraphics[width=\linewidth]{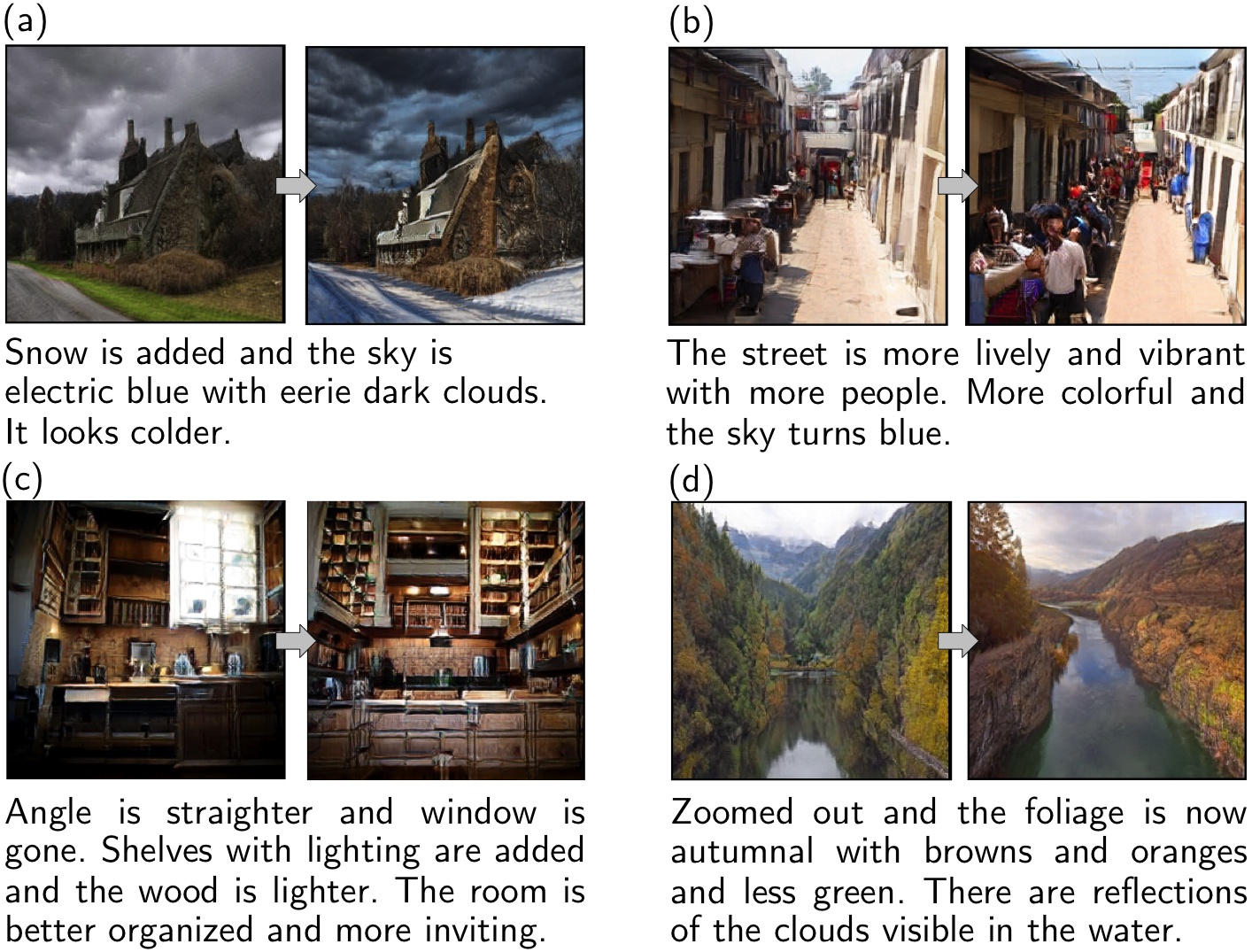}
\setlength{\belowcaptionskip}{-20pt}
\vspace{-10px}
\caption{Sample transformations and AMT annotations from all four image classes: (a) cottage, (b) medina, (c) kitchen, (d) lake.}
\label{fig:dirannotation}
\end{center}
\end{figure}

For each direction, we synthesize images in four classes within BigGAN-Places: \verb cottage ,  \verb kitchen ,  \verb lake , and \verb medina  (outdoor marketplace). These represent familiar visual scenes that balance indoor and outdoor, natural and built environments.
Direction annotations are collected using Amazon Mechanical Turk (AMT). Participants see a single image pair $[G(\mathbf{z}), G(\mathbf{z}+\alpha \mathbf{d}_{j})]$ and are asked to describe the main visual changes in composition and style between the two images, for a total of 5,120 annotations. Figure~\ref{fig:dirannotation} shows example images sequences and annotations. We provide further details on the AMT setup in Section S.1.

\vspace{-10px} \paragraph{Data normalization and post-processing.}

To clean and normalize the direction annotations produced in \cref{sec:annotation}, we first preprocess and lemmatize the labels using methods described in the supplement. Next, we post-process the labels by detecting phrases capturing decrease in a concept (e.g. \textit{less green}, or \textit{window is removed}), and assign them to individual negative directions. The result is a compact set of terms for human visual concepts describing each direction, which we refer to as \emph{cleaned annotations}. For example, the cleaned annotation for the direction shown in Figure~\ref{fig:dirannotation}a would read \textit{``snow, sky, electric, blue, eerie, dark, cloud, cold."} Across all classes, 2800 unique concepts appeared, 1372 repeated at least once.  122 appeared in all four classes. The number of distinct concepts used in each class independently is shown in Table~\ref{tab:concepts}. Of concepts that appeared more than 20 times in the entire dataset (across all 4 classes), 32\% are objects (e.g. \textit{cabinet, tree}), 48\% are attributes (e.g. \textit{warmer, brighter}), and 20\% describe scene- and object-level geometry (e.g. \textit{background, angle}). We provide a more detailed description of these categories and a breakdown of concepts by image class in Section S.2. 

The cleaned annotations indicate visual concepts that describe each of the LSDs. However, they do not isolate dimensions of variation that correspond to individual concepts; one direction may be described by multiple terms. 
To understand which visually salient terms can be mapped onto individual dimensions in the GAN’s representation, in Section \ref{sec:Distilled} we disentangle the annotated directions into a set of principal perceptual components in the $\mathcal{Z}$ latent space. 

\begin{table}[t!]
    \centering
    \normalsize
    \begin{tabular}{lccc}
        \toprule
        \phantom{ }                       & \textbf{Distinct} & \textbf{Repeated} & \textbf{Unique to}\\
        \textbf{Image class}\phantom{her} & \textbf{concepts} & \textbf{n$>$1 times} & \textbf{one class} \\
        \midrule
        \textbf{Cottage} & 1166 & 508  & 147 \\
        \textbf{Kitchen} & 1045 & 445  & 167 \\
        \textbf{Lake}    & 1167 & 479  & 153 \\
        \textbf{Medina}  & 1087 & 460  & 142 \\
        \textbf{\emph{All four}} & 2800 & 1372 & 609 \\
        \bottomrule
    \end{tabular}
    \caption{Distinct terms for concepts used in cleaned annotations, by class.  We focus on those repeated in multiple labels, of which just under half (44\%) appear in only one class.}
    \label{tab:concepts}
\end{table}

\paragraph{Evaluating direction quality.} While our main contribution is a procedure (Section \ref{sec:Distilled}) for extracting a set of disentangled, human-recognizable concepts from \emph{any} corpus of direction annotations, the method we describe for obtaining an initial set of directions also has advantages over related methods. To validate our decision to use the LSDs for annotation, we directly compared the annotations of LSDs in our dataset to two baselines: directions generated using the GANSpace method \cite{harkonen2020ganspace}, and randomly generated directions. For a subset of 600 LSDs (150 in each of four image classes), we collected 10 annotations per direction using the AMT protocol described in Section \ref{sec:annotation}. Additionally, we followed \cite{harkonen2020ganspace} and identified the same number of latent directions corresponding to principal components of feature tensors on the first three layers of $G$. Finally, we sampled 600 random directions of fixed magnitude. All directions were normalized and added to the same set of $\mathbf{z}$ with $\alpha = 6$.

Table \ref{tab:bleu} shows the results of our comparison. We find that LSDs elicit a more \textit{diverse} vocabulary of  both single-word concepts and their compositions. Additionally we measure inter-annotator BLEU \cite{papineni2002bleu} and inter-annotator BERTScore, where the latter leverages a large pretrained language model to measure semantic similarity between annotations \cite{zhang2020bertscore}. While our LSDs obtain lower inter-annotator BLEU scores than the baselines, they obtain a larger BERTScore, suggesting there is less lexical overlap but greater semantic overlap in how annotators describe LSDs compared to the baselines. This hypothesis is further substantiated by the greater diversity of $n$-grams in LSD annotations.

\begin{table}[h!]
    \centering
    \footnotesize
    \setlength{\tabcolsep}{3.7pt}
    \begin{tabular}{lccccc}
        \toprule
        \textbf{Directions} &  \textbf{1-grams} & \textbf{2-grams} & \textbf{3-grams} & \textbf{BLEU} & \textbf{BERTScore-R} \\
        \midrule
        Random & 2,316 & 14,913 & 22,938 & 8.86 & 0.375 \\
        GANSpace & 2,975 & 18,622 & 26,466 & 8.24 & 0.343 \\
        LSD (Ours) & 3,156 & 20,986 & 31,307 & 7.17 &  0.393 \\
        \bottomrule
   
    \end{tabular}
    \vspace{-5px}
    \caption{Comparison of diversity and reliability measures for 6000 annotations of directions added to the same set of $\mathbf{z}$. For the LSDs, observers recognize the most semantically similar changes per direction, and overall produce a larger number of both single-word concepts and their compositions.}
    \label{tab:bleu}
\end{table}

\subsection{Distilling directions for visual concepts}
\label{sec:Distilled}

Our goal is to identify a vocabulary of \emph{primitive} visual concepts, but as shown in Figure~\ref{fig:dirannotation}, the LSD annotations describe complex, compositional image changes, even after restricting annotation to layer-selective directions. We hypothesize that each annotated direction can be reconstructed from a set of \emph{distilled directions} associated with individual concepts in the annotation. In other words, 
\begin{equation}
    d(\textit{tall red building}) \approx d(\textit{tall}) + d(\textit{red}) + d(\textit{building})    
\end{equation}
This is a simplifying assumption (\emph{red} suggests a different color in in \emph{red hair} vs \emph{red brick}) \cite{Heim1998-HEISIG}. However, it provides a convenient (and empirically effective) mathematical framework for distilling directions for primitive concepts from compositional annotations. In particular, we can formulate learning of the visual concept vocabulary as a regularized linear regression of the form:
\begin{equation} \label{eq3}
\argmin_\mathbf{E} \left \| \mathbf{WE} - \mathbf{D} \right \|_F^2 + \lambda\left \| \mathbf{E} \right \|_F^2 
\end{equation}
where rows $i$ of word matrix $\mathbf{W}$ correspond to annotations, and columns $j$ of $\mathbf{W}$ to individual words. $\mathbf{W}_{i,j} = 1$ if word $i$ appears in cleaned annotation $j$. 
$\mathbf{WE}$ is thus a matrix of annotation embeddings that we can compare to $\mathbf{D}$, where rows $\mathbf{d}_{i}$ are the annotated directions in $\mathcal{Z}$ latent space. 

We may then solve analytically for $\mathbf{E}$: 
\begin{equation} \label{eq4}
\mathbf{E} = (\mathbf{W}^\top\mathbf{W} + \lambda \mathbf{I} )^{-1}\mathbf{W}^\top\mathbf{D}  
\end{equation}
where $\mathbf{I}$ is the identity matrix with the same size as $\mathbf{W}^\top\mathbf{W}$. 
The hyper-parameter $\lambda$ determines the balance between the L2 loss and the regularization of $\mathbf{E}$. 
We set $\lambda$ to 100 in our experiments. 

\begin{figure*}[t!]
\begin{center}
\includegraphics[width=.95\linewidth]{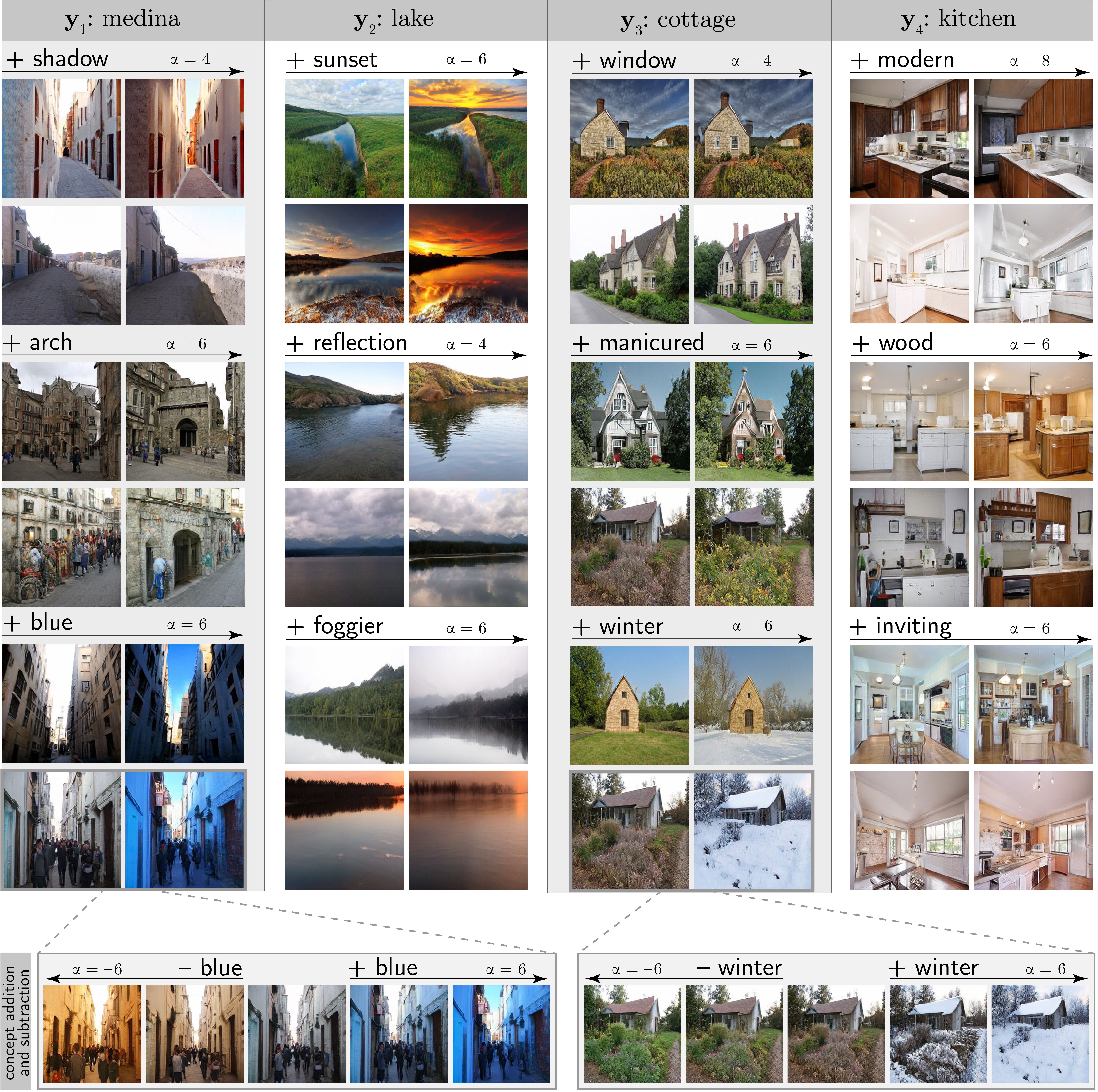}
\end{center}
    \vspace{-10px}
   \caption{Example visual concepts across four classes of visual scenes, each applied to two $\mathbf{z}$. This sample represents only 13 of 1372 unique concepts discovered in BigGAN-Places. Some concepts (such as \emph{blue}) occur in all scene classes. Others are characteristic of one or two (such as making a lake \emph{foggier} or a kitchen \emph{modern}). Bottom: in some cases, subtracting concepts can produce opposite transformations. For example, the subtraction of blue is the complementary orange, and the subtraction of winter is a spring scene. Additional examples with varying $\alpha$ are shown in the supplement.}
\label{fig:vconcepts}
\end{figure*}

The individual word embeddings $\mathbf{e}_j$ in the latent space of $G$ lie along the rows of $\mathbf{E}$. 
As in \cref{sec:directiongeneration},  transforming an image $G(\mathbf{z})$ along the distilled direction corresponding to concept $j$ is equivalent to moving in the direction $\mathbf{e}_j$ in $\mathcal{Z}$ latent space and passing the transformed $\mathbf{z}$ vector through the generator: $G(\mathbf{z} + \alpha\mathbf{e}_{j})$. 
The scaling parameter $\alpha$ determines the degree and type of transformation: a larger $\alpha$ introduces more of concept $j$ to $G(\mathbf{z})$, and in many cases, $-\alpha$ removes the visual concept from the scene. We note that the latent space is not perceptually uniform: steps of the same magnitude along different directions do not necessarily reflect the same amount of perceptual change. Continued work might map how this perceptual sensitivity to movement in each direction varies across the latent space.

Figures~\ref{fig:festive} and \ref{fig:vconcepts} illustrate the efficacy of applying our method to BigGAN-Places to disentangle directions corresponding to individual concepts, where each concept is associated with multiple annotated directions. We also tested the generalization of this approach to BigGAN-Imagenet, and show results in the supplement. Interestingly, \verb lake  is the only image class shared by both ImageNet and Places. For  the  same number of annotated directions (1280), the number of distinct concepts in the lake class for BigGAN-ImageNet is $<75\%$ of the number of distinct concepts in the BigGAN-Places lake class. This could reflect less scene diversity in comparable ImageNet classes due to less training data. Given that our method is generalizable and fairly model-agnostic, we suggest that it could be used in such a manner to characterize a given generator by the projection of concepts salient to humans into the set of concepts the model has learned.

\section{Evaluating distilled visual concepts}
\label{sec:Experiments}

We have now distilled our LSDs into a vocabulary of primitive visual concepts, each consisting of a short langauge description, \eg \emph{snow} or \emph{festive}, and an associated latent direction. Our next step is to evaluate how well the directions produce transformations in generated images that are faithful to their description. In other words, how often does adding the \emph{trees} direction to a starting representation clearly add trees to the image?

We study this empirically by conducting a series of human experiments in which crowdworkers are asked to discriminate which among several image transformations corresponds to a specific visual concept. One of the transformed images is constructed by adding the corresponding direction $\mathbf{d}$ to the starting $\mathbf{z}$, while the others are constructed by adding \emph{different directions from the vocabulary}. If humans reliably can discriminate which transformed image corresponds to the visual concept, that would suggest that the direction is faithful. The following three experiments adopt this structure and vary how the vocabulary is constructed in order to study different properties of the distilled directions. The first two experiments focus on whether the directions \textbf{generalize} across starting representations (\cref{sec:generalize-within}) and image classes (\cref{sec:generalize-across}). The final experiment explores whether they reliably \textbf{compose} with one another, supporting combinatorial extensions to the vocabulary (\cref{sec:composition}).

\subsection{Do concepts generalize across $\mathcal{Z}$?}
\label{sec:generalize-within}

We begin by asking whether distilled directions generalize to produce faithful transformations when added to unseen $\mathbf{z} \in \mathcal{Z}$, keeping all other inputs the same. Here, we fix a class $\mathbf{y}$ and \emph{only} vary the initial representation $\mathbf{z}$. This means that when we distill the vocabulary using Equation \ref{eq4}, we construct $\mathbf{W}$ using only annotations for which the human annotator saw images generated with the class $\mathbf{y}$.

For each visual concept $c_*$ and its distilled direction $\mathbf{d_*}$, we sample a $\mathbf{z} \in \mathcal{Z}$ and three distractor directions $\{\mathbf{d_1}, \mathbf{d_2}, \mathbf{d_3}\}$ from the remaining directions in the vocabulary. Human participants are shown an initial image $G(\mathbf{z}; \mathbf{y})$ and four transformed images $G(\mathbf{z} + \alpha\mathbf{d}_i; \mathbf{y})$ for $i = 1, 2, 3, *$ and are asked to discriminate which transformed image corresponds to $c_*$. If the direction $\mathbf{d_*}$ successfully generalizes to the new $\mathbf{z}$, then participants should reliably choose the image change generated by that direction.

We recruit crowdworkers from Amazon Mechanical Turk; full details about the AMT setup and other hyperparameters can be found in the supplement. To denoise, we generate three sets of $\mathbf{z}$s and distractors per concept in the vocabulary, and additionally show each $(\mathbf{z}$, $\mathbf{d})$ pair to five distinct participants, totaling 15 AMT HITs per concept.

\begin{figure}[t!]
    \centering
    \includegraphics[width=\linewidth]{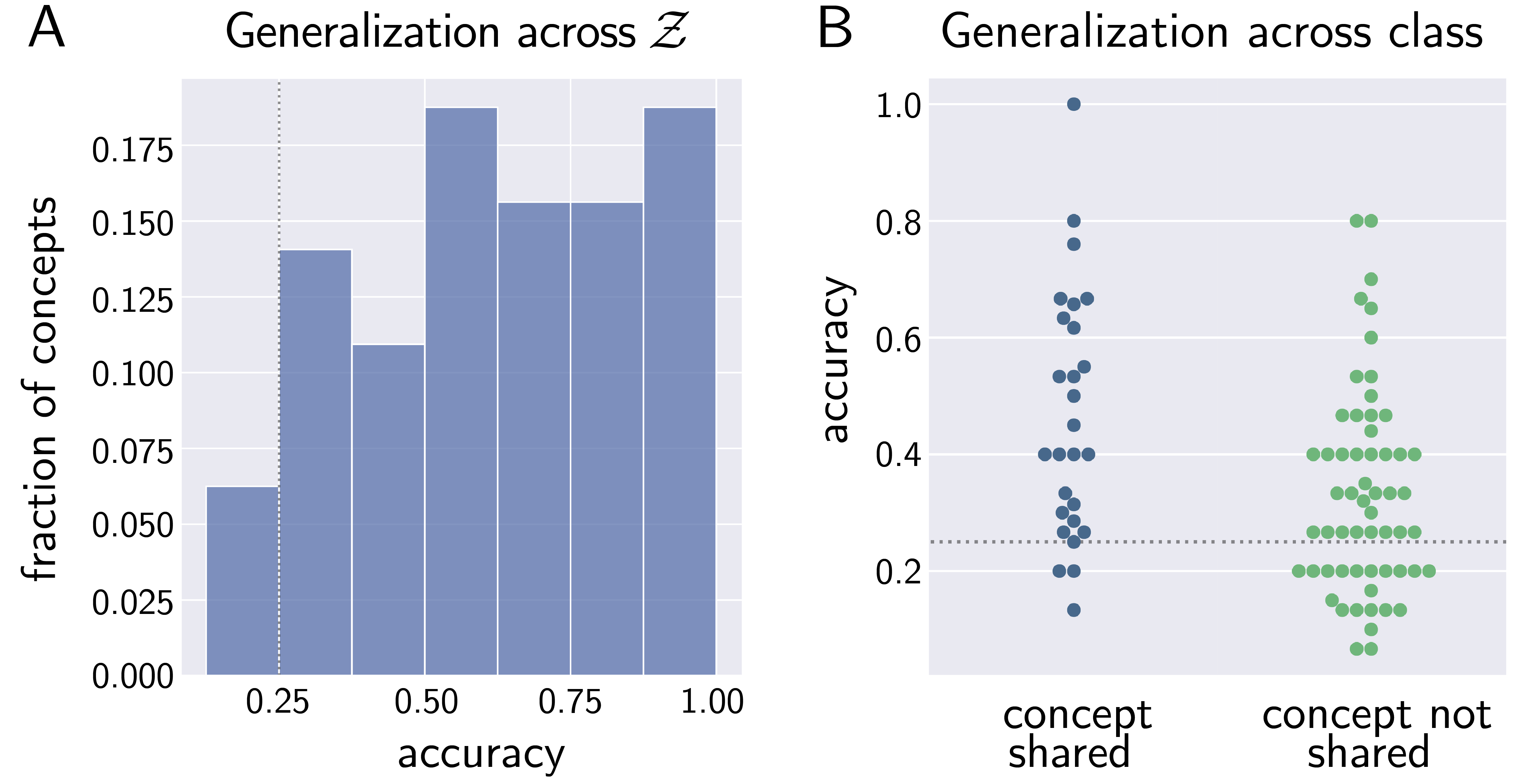}
    \vspace{-16px}
    \setlength{\belowcaptionskip}{-8pt}
    \caption{(a) Histogram from \cref{sec:generalize-within}, where \emph{accuracy} refers to the fraction of times that humans correctly recognized a specific concept. Dotted vertical line demarcates accuracy of random guessing. For 94\% of concepts, participants recognize the correct change more often than if they guessed randomly, suggesting that the directions generalize across $\mathbf{z}$. (b) Concept accuracies from the cross-class evaluation of \cref{sec:generalize-across}, bucketed by whether the concept appeared in both annotations for the training class and the test class. Some concepts (typically objects and attributes) exhibit strong cross-class generalization, with one being correctly recognized by every observer. Other concepts fail to generalize even when they appear in annotations for both classes, suggesting BigGAN has not entirely disentangled concept from class.}
    \label{fig:z_cat_gen}
\end{figure}

\paragraph{Distilled directions generalize to novel inputs.} \cref{tab:accuracies} shows human accuracies by image class. Participants identify the correct image transformation more than 60\% of the time, providing strong evidence that the distilled directions generalize across the representation space. Figure \ref{fig:z_cat_gen}a shows that many concepts are recognized with higher accuracy than reported in \cref{tab:accuracies}, and only about 6\% of concepts are recognized at the level of chance. \textit{Attributes} are the most likely category of concepts to be accurately detected (75\%). We include a further breakdown by concept in Section S.3.  

\paragraph{Detecting concepts with an SVM.}
We replicated Experiment 1 using a linear classifer to detect concepts added to generated images, providing additional evidence that our vocabulary generalizes across $\mathcal{Z}$. For each of the the 20 most frequent concepts in all four classes, we trained a linear SVM to distinguish the addition of that concept to an image from the addition of a randomly sampled distractor, and tested on held out images. Mean classification accuracy was significantly above chance in all classes (\textit{cottage}: 80.2\%, \textit{kitchen}: 73.4\%, \textit{lake}: 79\%, \textit{medina}: 77.3\%), and like humans, accuracy was highest overall for \textit{attributes} (82.8\%). We provide a per concept breakdown in the supplement.

\begin{table}[t!]
    \centering
    \footnotesize
    \begin{tabular}{lccccc}
        \toprule
        \textbf{Experiment} & \textbf{Kitchen} & \textbf{Lake} & \textbf{Medina} & \textbf{Cottage} & \textbf{Avg.} \\
        \midrule
        Generalize $\textbf{z}$  & .60 & .76 & .62 & .64 & .66 \\
        Generalize $\textbf{y}$ & .37 & .39 & .43 & .37 & .39 \\
        Composition & .40 & .44 & .51 & .41 & .44 \\
        \bottomrule
    \end{tabular}
    \vspace{-3px}
    \caption{Human accuracy discriminating a target concept from three distractors, where the concept is visualized by applying its associated direction to new $\mathbf{z}$ (Generalize $\textbf{z}$), new classes (Generalize $\textbf{y}$), and compositions of directions (Composition).}
    \label{tab:accuracies}
\end{table}

\subsection{Do concepts generalize across classes?} 
\label{sec:generalize-across}

Visual concepts are context-sensitive. For example, making a kitchen scene \emph{brighter} might involve adding additional light fixtures, while making a cottage scene brighter will likely involve intensifying the sun. Despite the differences between these image transformations, both are instantiations of the visual concept \emph{brighter}. At the same time, some visual concepts might be unique to a context. The kitchen class exclusively features  concepts like \emph{cabinets} and \emph{appliances}, while the lake class features \emph{snow} and \emph{mountains}. This raises the question: if we construct a vocabulary using annotations from one image class, do the resulting directions produce faithful transformations on other classes?

\begin{figure}[H]
\begin{center}
\includegraphics[width=\linewidth]{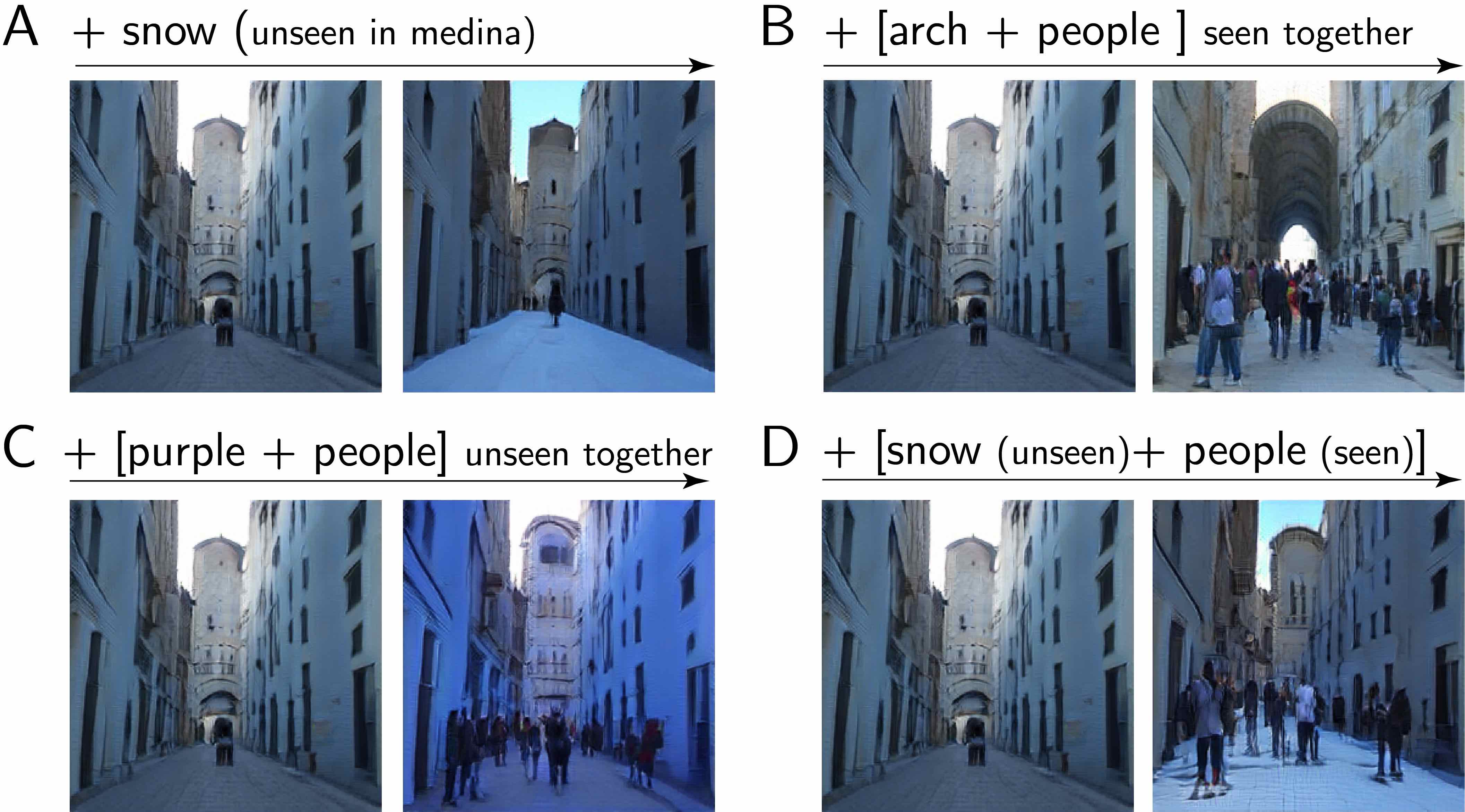}
\setlength{\belowcaptionskip}{-10pt}
\caption{Several image changes produced by decomposed directions applied to the same starting image of a medina. Directions generalize (a) across and (b) within class. Two directions can be composed regardless of whether the corresponding concepts (c) did not co-occurr in the original corpus or (d) did co-occur.}
\label{fig:dircomposition}
\end{center}
\end{figure}

We now repeat our evaluation from \cref{sec:generalize-within}, but instead of fixing $\mathbf{y}$ in the evaluation, we choose it at random from the set of classes not used to construct $\mathbf{E}$ in Equation \ref{eq4}. Hence, when evaluating the \texttt{kitchen} vocabulary, we generate images and transformations with the \texttt{lake}, \texttt{cottage}, or \texttt{medina} class. We draw several conclusions.
\vspace{-20px}
\paragraph{Generalization across class is most robust when concepts are shared between classes.} Figure \ref{fig:z_cat_gen}b shows that participants recognize concepts most often when the concept appears in the vocabulary for both classes. This agrees with the intuition that it should be difficult to add a visual concept to image when that concept is foreign to the context, \eg adding \emph{appliances} to a lake scene. For these transformations to suceed, BigGAN would have to generate out of distribution images.
\vspace{-8px}
\paragraph{However, distilled directions still generalize across classes.}  
Even though cross-class generalization is harder than within-class generalization, humans still recognize the target visual concept a majority of the time. This even includes some out-of-distribution generalizations like the one shown in Figure \ref{fig:dircomposition}, which inserts snow into a \texttt{medina}, despite snow being unseen in \texttt{medina} training images.

\subsection{Do concepts compose?}
\label{sec:composition}

In the previous experiments, our vocabulary consisted of primitive visual concepts such as \emph{mountain} and \emph{dark}. Can we construct more complex visual concepts from these primitive ones? One way to do this would be to compose the primitive concepts conjunctively: given a \emph{mountain} direction and a \emph{dark} direction, construct a $\textit{mountain} \wedge \textit{dark}$ by simply averaging the two directions.

Our goal in this section evaluate how often composition of this kind succeeds. We repeat the evaluation from \cref{sec:generalize-within}, now constructing the vocabulary by conjunctively composing every pair of primitive concepts from the original vocabulary. Formally, given a primitive vocabulary $V$ for a fixed concept $\mathbf{y}$ and two directions $\mathbf{a}, \mathbf{b} \in V$, we define their composition $\mathbf{a} \circ \mathbf{b}$ to be $(\mathbf{a} + \mathbf{b}) / 2$ and define our new vocabulary to be $V' = \{\mathbf{a} \circ \mathbf{b} : (\mathbf{a}, \mathbf{b}) \in V^2\}$. In practice, $V'$ is quite large because it has quadratically many concepts, so we select a random subset of 50 compositions.

As before, for each direction $\mathbf{a} \circ \mathbf{b} \in V'$, we sample a representation $\mathbf{z} \in \mathcal{Z}$ and three distractor directions. However, now we choose two of the distractors to be compositions of $\mathbf{a}$ and $\mathbf{b}$ with other primitives. Specifically, we sample two additional directions $\mathbf{c}, \mathbf{d} \in V - \{\mathbf{a}, \mathbf{b}\}$ and use $\mathbf{a} \circ \mathbf{c}$, $\mathbf{b} \circ \mathbf{d}$, and $\mathbf{c} \circ \mathbf{d}$ as distractors. Participants then discriminate which transformed image contains both $\mathbf{a}$ and $\mathbf{b}$.

\begin{figure}[t!]
    \centering
    \includegraphics[width=\linewidth]{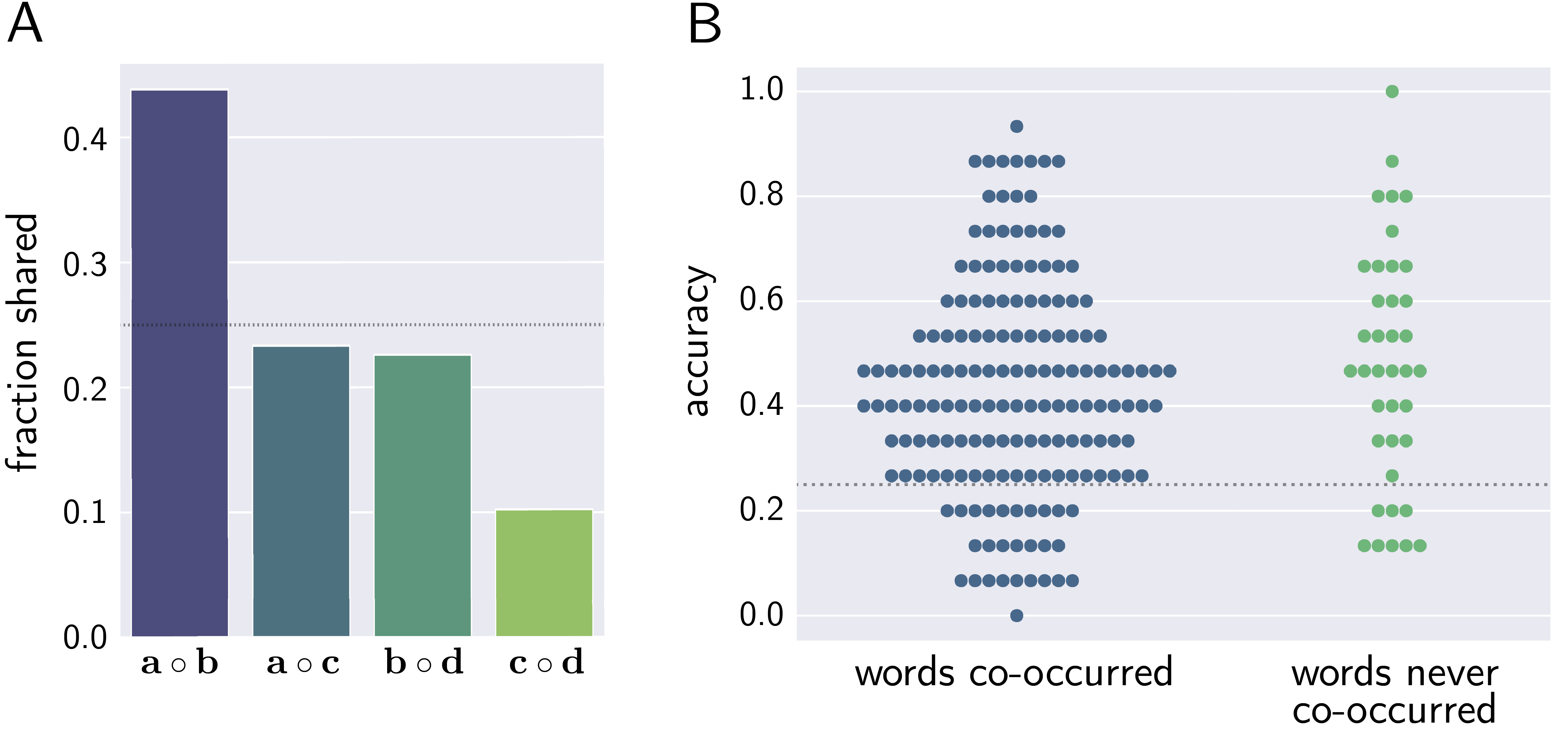}
    \vspace{-10px}
    \setlength{\belowcaptionskip}{-10pt}
    \caption{(a) Fraction of times humans chose each composition in \cref{sec:composition}. $\mathbf{a}$ and $\mathbf{b}$ are target directions, while $\mathbf{c}$ and $\mathbf{d}$ are randomly chosen distractors. Observers frequently recognize the correct composition, but even when not, they prefer partially correct compositions, suggesting the decomposed directions compose faithfully. (b) Fraction of times humans recognized each concept composition, bucketed by whether composed concepts co-occurred in the original corpus. Both classes of composition perform have comparable mean accuracies, suggesting many of the directions in the vocabulary can be faithfully composed.}
    \label{fig:composition}
\end{figure}
\vspace{-8px}
\paragraph{Distilled directions compose to produce new and recognizable concepts.} Even though compositional changes are harder to discriminate, participants still predict the correct change reliably above chance. Furthermore, Figure \ref{fig:composition}a shows that when participants choose a distractor, they tend to pick distractors closest to the target, i.e. $\mathbf{a} \circ \mathbf{c}$ or $\mathbf{b} \circ \mathbf{d}$.
\vspace{-8px}
\paragraph{Composition produces faithful transformations even when concepts did not co-occur in the training data.} Figure \ref{fig:composition}b shows that participants recognize composed concepts regardless of whether the constituent concepts ever appeared together in a single LSD description. Figure \ref{fig:dircomposition} shows an example, in which the \emph{purple} and \emph{people} concepts (unseen together during training) can be composed to produce an image of a purple \texttt{medina} filled with people.


\section{Conclusion}

We introduce a new procedure for building open-ended vocabularies of primitive visual concepts represented in GANs’ latent spaces, and show that these concepts are reliably recognizable and freely composable. This work represents an important step toward bridging the representational gap between human perception and artificial generators. Future work could explore the use of our approach with generators other than BigGAN, such as StyleGAN. 

\vspace{10px}
\noindent{\small\bf{Acknowledgements.}} {\small We thank the MIT-IBM Watson AI Lab for support, and IBM for the donation of the Satori supercomputer that enabled training BigGAN on MIT Places. We also thank Luke Hewitt for valuable discussion and insight.}

\newpage\clearpage

{\small
\bibliographystyle{ieee_fullname}
\bibliography{main.bib}
}

\newpage\clearpage

\end{document}


\vspace{-20mm}
\title{Supplemental Materials for \\ 
Toward a Visual Concept Vocabulary for GAN Latent Space}

\maketitle
\ificcvfinal\thispagestyle{empty}\fi

\renewcommand{\thefigure}{S.\arabic{figure}}
\setcounter{figure}{0}
\renewcommand{\thetable}{S.\arabic{table}}
\setcounter{table}{0}
\renewcommand{\thesection}{S.\arabic{section}}
\setcounter{section}{0}

\section{Annotation collection and processing} 
\label{sec:collection}

\begin{figure}[b!]
\begin{center}
   \includegraphics[width=\linewidth]{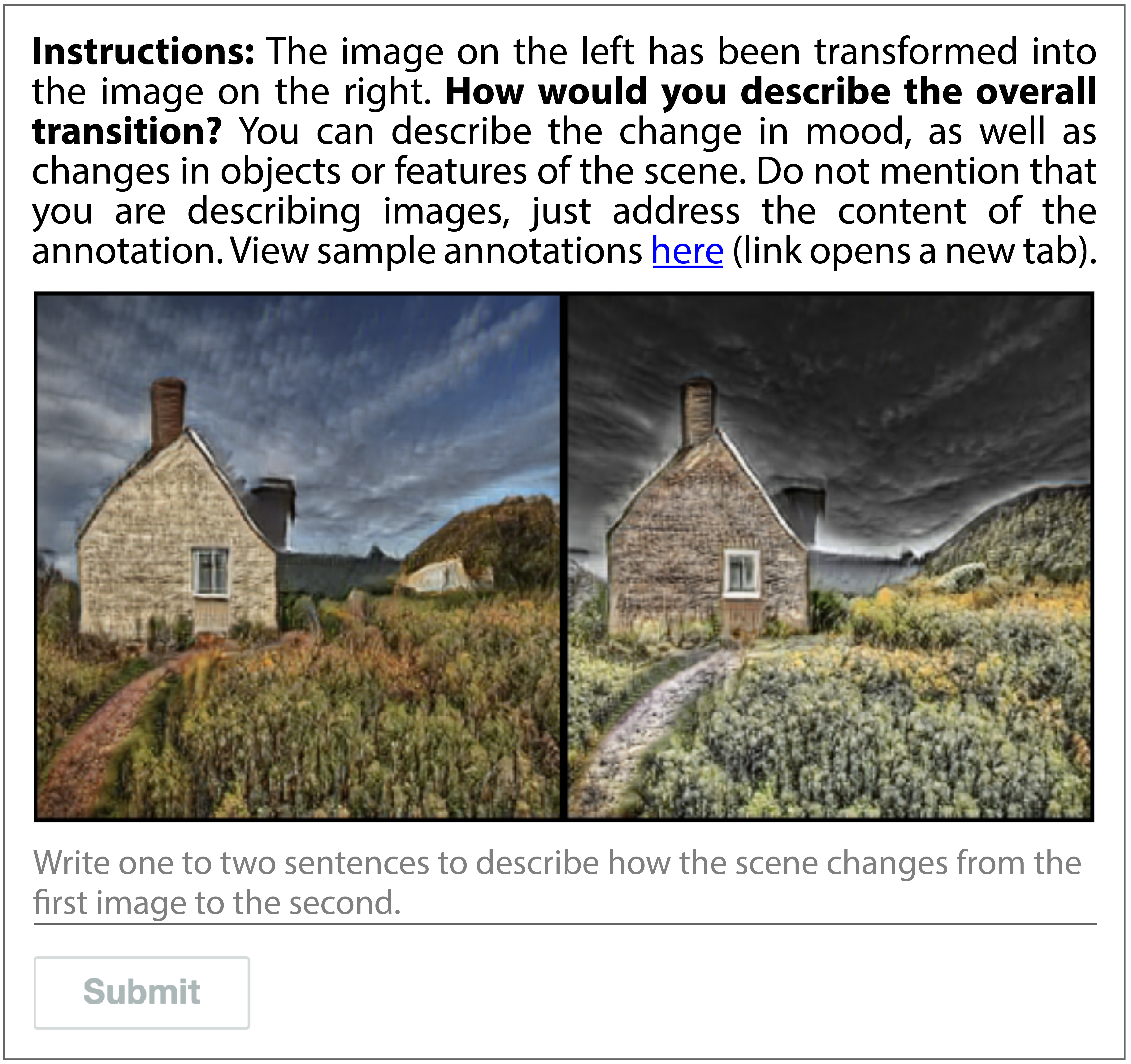}
\end{center}
   \vspace{-10px}    
   \captionsetup{}
   \caption{Example annotation HIT. Annotators are shown $G(\mathbf{z}_i;\mathbf{y})$ (left) and $G(\mathbf{z}_i + \alpha\mathbf{d}_{i,j};\mathbf{y})$ (right) and asked to write freeform text to describe the change from L to R. We used a value of $\alpha = 6$ for all experiments.}
\label{fig:annotationex}
\end{figure}

\paragraph{Collection} As described in Section 3.2, we collect annotations using Amazon Mechanical Turk (AMT) for layer-selective directions visualized in four classes (cottage, kitchen, lake, medina). Instructions and an example task are found in  Figure~\ref{fig:annotationex}. 
We require workers to be located in the U.S., with $>97\%$ HIT acceptance rate and $>100$ HITs accepted. Workers were paid $\$0.06$ per annotation. 

\vspace{-12px}
\paragraph{Normalization and post-processing} We normalize direction annotations before applying the method described in Section 3.3 to decompose them into a vocabulary of primitive visual concepts. We use \emph{pyspellchecker} to automate simple corrections, keeping the original string if there is no word with an edit distance less than 3. Lemmatizing is done with NLTK WordNetLemmatizer, discarding common terms used to describe the setting (e.g., \emph{image}, \emph{scene}) or image class (e.g., \emph{house}, \emph{lake}).  We then run a basic sentiment analysis script to detect modifier words indicating whether a concept is being added (e.g., \emph{appears, added, more}) or taken away (e.g., \emph{disappears, removed, less}, \emph{goes from}) in an image transformation. This simple approach worked sufficiently well to disambiguate different uses and positive vs. negative sentiment of a concept.

\section{Concept categorization}
\label{sec:categories}

Here we provide a breakdown of concepts into three categories reflecting their use, as described in Section 3.2 of the main paper. All concepts that appeared more than five times in each image class were categorized by the authors, as well as all concepts that appeared more than 20 times across all four classes. We sort concepts into three broad categories: \textit{object}, including collective nouns and regions of scenes (e.g., \emph{people, ocean, road}), \textit{attribute} (descriptors of object and scene qualities, including color), and \textit{geometry} (scene- and object-level geometry, including size, perspective, and position). We report results in Table~\ref{tab:category}.

Attributes are the largest category of concepts in every image class. Concepts describing color and light make up 50\% of all attributes: 38\% are chromatic color, 12\% are related to light and dark. Attributes are also the most reliably detected concepts, both automatically and by humans (see Section~\ref{sec:accuracies}).

\begin{table}[b!]
    \centering
    \footnotesize
    \setlength{\tabcolsep}{6pt}
    \begin{tabular}{rccccc}
        \toprule
           &  \textbf{Cottage} & \textbf{Kitchen} & \textbf{Lake} & \textbf{Medina} &  \textbf{All classes} \\
        \midrule
        \textbf{Object}    & 35\% & 29\% & 24\% & 25\% & 32\% \\
        \textbf{Attribute} & 48\% & 50\% & 54\% & 54\% & 48\% \\
        \textbf{Geometry}  & 17\% & 21\% & 22\% & 21\% & 20\% \\
        \midrule
        \textbf{ \# of terms} & 178 & 140 & 184 & 139 & 152 \\
        \bottomrule
    \end{tabular}
    \vspace{-5px}
    \captionsetup{}
    \caption{Distribution of frequently used concepts across three classes: names of objects, scene- and object-level geometry, and other attributes (such as color or lighting).  This shows terms that appear 5+ times within each class, and 20+ times across all classes.}

    \label{tab:category}
\end{table}

\vspace{12pt}

\vspace{-8px}
\section{Concept detection accuracies}
\label{sec:accuracies}

Here we report human and SVM accuracies in detecting the addition of individual concepts to generated images.

\paragraph{Human performance per concept.} Experiment 1 (described in Section 4.1 of the main paper) evaluated the generalizability of our vocabulary across $\mathcal{Z}$ by measuring human accuracy discriminating a target concept among three distractors. Table~\ref{tab:hume} reports mean accuracy across 15 workers per concept, for the 20 most frequent concepts in each class. Mean human accuracy classifying \textit{attributes} (0.79, $\sigma$ = .21) is higher than either objects (0.64, $\sigma$ = .25 ) or geometry (0.47, $\sigma$ = 0.17). All but one of the attributes shown in Table~\ref{tab:hume} describe chromatic color or light, which we might expect to be more reliably discriminable across images and observers. 

\begin{table}[t!]
    \centering
    \footnotesize
    \setlength{\tabcolsep}{2pt}
    \begin{tabular}{rl@{\hskip .1in}rl@{\hskip .1in}rl@{\hskip .1in}rl}
        \toprule
        \multicolumn{2}{c}{\textbf{\hspace{8pt}Cottage}}  & \multicolumn{2}{c}{\textbf{\hspace{3pt}Kitchen}}  & \multicolumn{2}{c}{\textbf{\hspace{16pt}Lake}} & \multicolumn{2}{c}{\textbf{\hspace{19pt}Medina}}   \\
        \midrule
        tree & 0.80 & wall & 0.13 & water & 0.53 & alley & 0.60  \\
        \blue{color} & 0.20 & window & 0.87 & tree & 0.93 & wall & 0.60  \\
        sky & 0.53 & cabinet & 0.60 & sky & 0.73 & people & 0.73  \\
        building & 0.20 & \blue{color} & 0.67 & cloud & 0.73 & \blue{color} & 0.67  \\
        grass & 0.80  & \blue{white} & 0.87 & \blue{color} & 0.87 & \blue{darker} & 0.93  \\
        \blue{green} & 1.00 & \blue{lighter} & 0.80 & \blue{blue} & 1.00 & street & 0.20  \\
        window  & 0.73 & counter & 0.53 & \blue{darker} & 0.93 & \blue{blue}  &  0.87 \\
        \blue{darker} &  0.73 & \blue{darker} & 0.93 & \blue{green} & 0.87 & sky  &  0.40 \\
        roof & 0.40 & \blue{brown} & 1.00 & reflection & 0.80 & window &  0.67 \\
        \blue{white} & 0.87 & \blue{brighter} & 0.73 & mountain & 1.00 & \blue{light}  &  0.60 \\
        \green{front} &  0.67 & \blue{wood} & 0.87 & land & 0.73 & \blue{brighter}  &  0.53 \\
        \blue{red} & 0.67 & floor & 0.33 & \blue{brighter} & 0.47 & door  &  0.33 \\
        \green{smaller} & 0.53 & \green{space} & 0.27 & grass & 0.87 & \blue{white}  &  0.87 \\
        snow &  0.93 & \blue{blue} & 1.00 & \green{background} & 0.33 & \blue{red}  &  0.80 \\
        \green{angle} & 0.53 & \blue{yellow} & 0.93 & \blue{lighter} & 0.40 & \blue{B\&W}  &  0.80 \\
        \blue{blue} & 0.80 & \green{smaller} & 0.53 & building & 1.00 & \blue{yellow}  &  1.00 \\
        \blue{B\&W} & 1.00 & \green{angle} & 0.13 & \blue{yellow} & 1.00 & arch  &  1.00 \\
        \green{larger} & 0.47 & \blue{warmer} & 0.73 & \blue{B\&W} & 0.93 & \green{background}  &  0.47 \\
        cloud & 0.53 & \blue{red} & 1.00 & \blue{day} & 0.13 & road  &  0.60 \\
        \blue{brown} & 0.67 & table & 0.13 & \blue{sunset} & 0.93 & \green{wider}  &  0.73 \\
        \midrule
        \textbf{Average}& 0.65  & & 0.65 & & 0.76 & & 0.67\\
        \bottomrule
   
    \end{tabular}
    \vspace{-5px}
    \captionsetup{}
    \caption{Human accuracy detecting the 20 most frequent concepts by category in Experiment 1. Chance is 0.25. Black concepts are objects (including collective nouns and larger scene regions, e.g. water), \blue{blue} concepts are attributes (adjectives, including colors), and \green{green} concepts describe scene- and object-level geometry.}
    \label{tab:hume}
    \vspace{-8pt}
\end{table}

\vspace{-8pt}
\paragraph{SVM performance per concept.} As described in Section 4.1, we replicated Experiment 1 using a linear SVM to distinguish the addition of a particular concept to images from the addition of distractors. For the top 20 most frequent concepts in each of the four classes, 64 $\mathbf{z}$ were randomly sampled, and two classes of images were created to train the SVM: $G(\mathbf{z}+\mathbf{d}_*,\mathbf{y})$ where $\mathbf{d}_*$ is the target concept, and $G(\mathbf{z}+\mathbf{d}_j,\mathbf{y})$ where the $\mathbf{d}_j$ are randomly sampled from the other 19 concepts. 20\% of images were held out for testing.

We report classification accuracy for the top 20 concepts in all four classes in Table~\ref{tab:svm}. The color of each concept reflects its category (see Section~\ref{sec:categories}). Like in the human experiment, mean SVM accuracy classifying \textit{attributes} (0.83, $\sigma$ = 0.09) is higher than either objects (0.75, $\sigma$ = 0.08) or geometry (0.72, $\sigma$ = 0.08).

\begin{table}[t!]
    \centering
    \footnotesize
    \setlength{\tabcolsep}{2pt}
    \begin{tabular}{rl@{\hskip .1in}rl@{\hskip .1in}rl@{\hskip .1in}rl}
        \toprule
        \multicolumn{2}{c}{\textbf{\hspace{8pt}Cottage}}  & \multicolumn{2}{c}{\textbf{\hspace{3pt}Kitchen}}  & \multicolumn{2}{c}{\textbf{\hspace{16pt}Lake}} & \multicolumn{2}{c}{\textbf{\hspace{19pt}Medina}}   \\
        \midrule
        tree & 0.77 & wall & 0.73 & water & 0.73 & alley & 0.73  \\
        \blue{color} & 0.92 & window & 0.81 & tree & 0.81 & wall & 0.73  \\
        sky & 0.77 & cabinet & 0.65 & sky & 0.65 & people & 0.73  \\
        building & 0.77 & \blue{color} & 0.85 & cloud & 0.85 & \blue{color} & 0.85  \\
        grass & 0.77  & \blue{white} & 0.62 & \blue{color} & 0.69 & \blue{darker} & 1.00  \\
        \blue{green} & 0.73 & \blue{lighter} & 0.88 & \blue{blue} & 0.96 & street & 0.77  \\
        window  & 0.77 & counter & 0.50 & \blue{darker} & 0.85 & \blue{blue}  &  0.92 \\
        \blue{darker} &  0.85 & \blue{darker} & 0.95 & \blue{green} & 0.95 & sky  &  0.65 \\
        roof & 0.73 & \blue{brown} & 0.80 & reflection & 0.85 & window &  0.77 \\
        \blue{white} & 0.73 & \blue{brighter} & 0.85 & mountain & 0.69 & \blue{light}  &  0.77 \\
        \green{front} &  0.81 & \blue{wood} & 0.81 & land & 0.62 & \blue{brighter}  &  0.85 \\
        \blue{red} & 0.85 & floor & 0.69 & \blue{brighter} & 0.85 & door  &  0.77 \\
        \green{smaller} & 0.73 & \green{space} & 0.58 & grass & 0.85 & \blue{white}  &  0.81 \\
        snow &  0.88 & \blue{blue} & 0.81 & \green{background} & 0.69 & \blue{red}  &  0.85 \\
        \green{angle} & 0.77 & \blue{yellow} & 0.92 & \blue{lighter} & 0.81 & \blue{B\&W}  &  0.62 \\
        \blue{blue} & 0.77 & \green{smaller} & 0.77 & building & 0.88 & \blue{yellow}  &  0.88 \\
        \blue{B\&W} & 0.88 & \green{angle} & 0.65 & \blue{yellow} & 0.65 & arch  &  0.62 \\
        \green{larger} & 0.85 & \blue{warmer} & 1.00 & \blue{B\&W} & 0.73 & \green{background}  &  0.62 \\
        cloud & 0.73 & \blue{red} & 0.65 & \blue{day} & 0.88 & road  &  0.77 \\
        \blue{brown} & 0.96 & table & 0.77 & \blue{sunset} & 0.85 & \green{wider}  &  0.77 \\
        \midrule
        \textbf{Average}& 0.80  & & 0.76 & & 0.79 & & 0.77\\
        \bottomrule
   
    \end{tabular}
    \vspace{-5px}
    \captionsetup{}
    \caption{SVM accuracy classifying 20 most frequent concepts by category. The same color scheme is used as in Table~\ref{tab:hume}. Chance is 0.50.}
    \label{tab:svm}
\end{table}


\section{Generalization to BigGAN-ImageNet}

\noindent Our method generalizes to BigGAN-ImageNet, as referenced in the main paper.  We include details in this section. The generalizability of our approach suggests that it could be used to characterize a given generator by the projection of concepts salient to humans into the set of concepts the model has learned.

\begin{figure*}
\begin{center}
   \includegraphics[width=\linewidth]{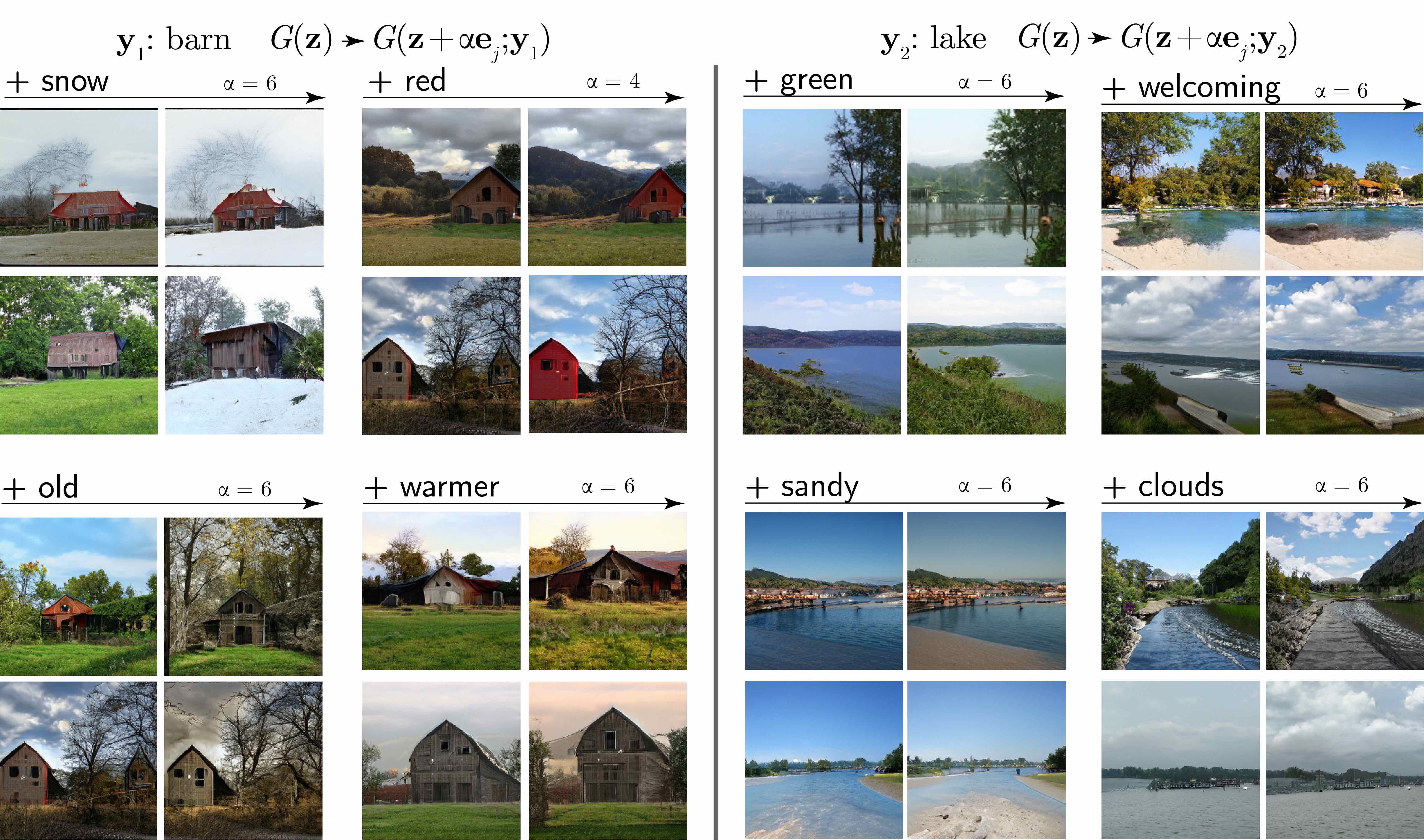}
\end{center}
   \captionsetup{}
   \caption{Example visual concepts found in the latent space of BigGAN-ImageNet using our method. The \emph{lake} class is the only visual scene class shared by both BigGAN-ImageNet and BigGAN-Places. For the same number of annotated directions (1280), the number of distinct concepts in the \emph{lake} class for BigGAN-ImageNet is $<75\%$ of the number of distinct concepts in the \emph{lake} class for BigGAN-Places (see Section 3.2, Table 1). This could reflect less scene diversity in comparable ImageNet classes due to less training data.}
\label{fig:Imagenetex}
\end{figure*}

\vspace{-8pt}
\paragraph{Distilling visual concepts.}  We generate layer-selective directions using the method described in Section 3.1 for 64 randomly selected $\mathbf{z}$ in two classes of BigGAN-ImageNet that best resemble classes of BigGAN-Places: \emph{lakes} (shared by both datasets) and \emph{barns} (similar to cottages). As in our procedure for BigGAN-Places, 1280 layer-selective directions are found in each class. Annotations are then collected on AMT, normalized, and post-processed using the procedure described in Sections 3.2 and \ref{sec:collection}. From the annotated layer-selective directions, we use the method described in Section 3.3 to distill visual concepts and associated directions in latent space. We find that 1198 unique terms are used to describe barns, with 555 repeated at least once, and 867 unique terms are used to describe lakes, with 390 repeated at least once. Selected directions in both classes are visualized in  Figure~\ref{fig:Imagenetex}.

\begin{figure}[H]
\begin{center}
   \includegraphics[width=\linewidth]{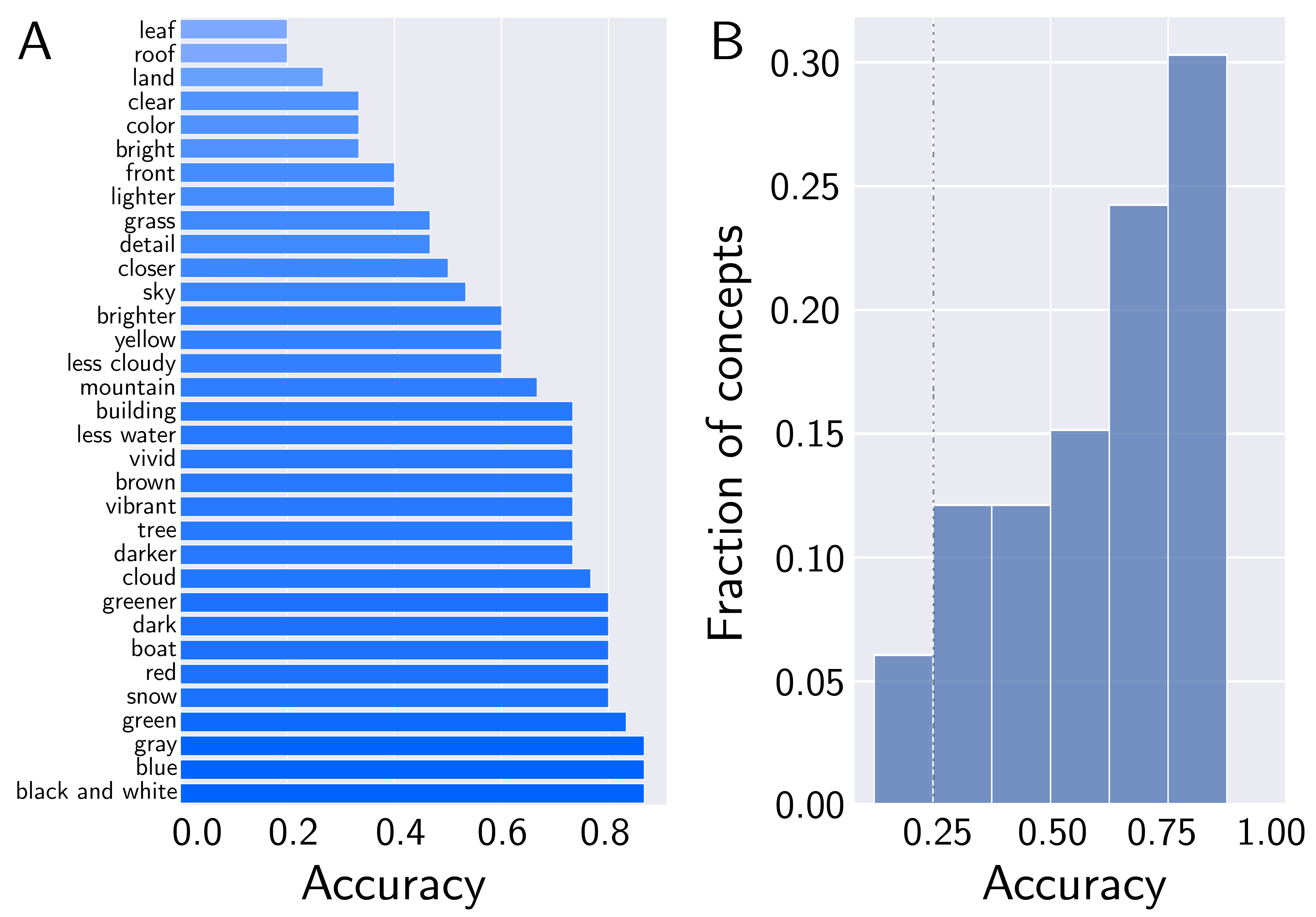}
\end{center}
   \captionsetup{}
   \caption{Task accuracies for concepts computed across $\mathbf{z}$, workers, and class. (a) Accuracies for concepts that appeared more than 20 times in the annotation dataset. Some concepts (including color changes like \emph{green, gray, blue, black and white}) are reliably recognized across most $\mathbf{z}$, while others (such as \emph{leaf} and \emph{roof}) are not recognized with accuracy above chance. (b) Histogram of concept accuracies across all concepts. The dotted vertical line shows the accuracy of random guessing (0.25).}
\label{fig:Imageneteval}
\end{figure}

\vspace{-8pt}
\paragraph{Concept evaluation.}
Following Section 4.1, we use a forced choice task on AMT to evaluate the salience of visual concepts in the latent space of BigGAN-ImageNet and their interpretability across different $\mathbf{z}_i$. Results across both classes are shown in Figure~\ref{fig:Imageneteval}.

Using the procedure described in Section 4.1 and visualized in Appendix C, AMT workers are recruited to identify each concept within a set of distractors. Specifically, for each concept $\mathbf{c_*}$ and its distilled direction $\mathbf{d}_*$, we sample a novel $\mathbf{z}$ from the $\mathcal{Z}$ latent space as well as three distractor directions $\{ \mathbf{d}_1, \mathbf{d}_2, \mathbf{d}_3\}$ sampled uniformly at random from the remaining directions. Workers are shown an initial image $G(\mathbf{z};\mathbf{y})$  and four modified images
$G(\mathbf{z}+ \alpha\mathbf{d}_i;\mathbf{y})$ for $i = 1; 2; 3; *$ and are asked to discriminate which modified image corresponds to $\mathbf{c}_*$. If the direction $\mathbf{d}_*$ successfully generalized to $\mathbf{z}$, then workers should reliably choose the image change generated by that direction. We run the evaluation on 3 $\mathbf{z}$ per direction and show each ($\mathbf{z}$, $\mathbf{d}$) pair to 5 distinct workers. We use $\alpha = 6$ in our experiments. We find that workers reliably choose the correct image with $61.1\%$ overall accuracy across ($\mathbf{z}$, $\mathbf{d}$) in the \emph{barn} class, and $61.6\%$ overall accuracy in the \emph{lake} class. Observers only fail to discriminate about 6\% of concepts. For all other concepts, observers recognize the correct change more often than if they guessed randomly, demonstrating that our method is successful at discovering directions that generalize across the latent space of BigGAN-ImageNet.

\begin{figure*}[t!]
\begin{center}
   \includegraphics[width=\linewidth]{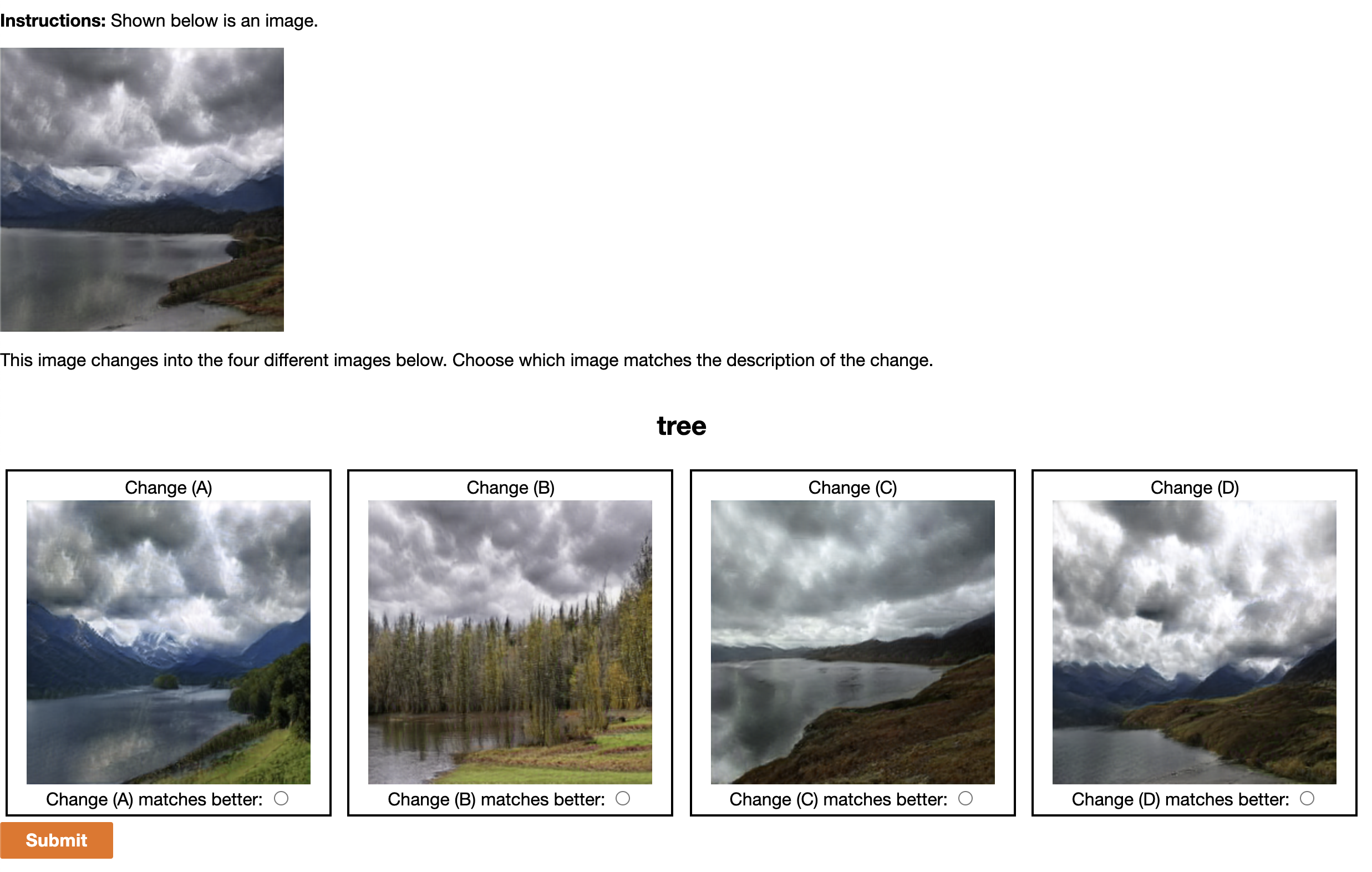}
\end{center}
   \captionsetup{}
   \caption{Example multiple choice HIT used to test concept generalization across image class. Here, the \textbf{tree} direction in $\mathcal{Z}$ latent space was learned in the \emph{cottage} class, and is being tested in the \emph{lake} class. The same multiple choice format is used to test concept composition, where a target composition (e.g. \textbf{tree, greener}) is described, and workers select which of four images best captures the composition.}
\label{fig:AMTex}
\vspace{-7pt}
\end{figure*}

\section{Generalization and composition experiments}

\paragraph{Experimental Paradigm.} In Figure~\ref{fig:AMTex} we show a screenshot of the paradigm used to collect data on AMT for analyses in Sections 4.1, 4.2, and 4.3, testing direction generalization across $\mathcal{Z}$ and image class, and composition. 

Workers are shown an original image $G(\mathbf{z};\mathbf{y})$ for randomly selected $\mathbf{z}$ and asked to identify which of four transformed images best corresponds to a named concept $\mathbf{c}_*$. $\mathbf{c}_*$ is randomly positioned among four distractors. Concepts used to create the distractor images are randomly sampled from the list of concepts that appeared more than five times in the annotation data. $\alpha$ was set to 6 for all experiments, and 5 distinct workers performed each task.

To test salience of concepts and their generalization across $\mathbf{z}$ in a given class (Section 4.1), concepts are tested for the same class $\mathbf{y}$ they were drawn from, for 3 different $\mathbf{z}$ per concept. To test generalization across class (Section 4.2), concepts are tested on a class (selected at random for each task) other than the one they were drawn from. Concept composition is tested using the method described in Section 4.3. Workers select between 4 modified images, one of which corresponds to a pair of concepts named in the task (e.g. [\textbf{tree, greener}]) and is randomly positioned among 3 other compositions (e.g. [\textbf{tree}, \textbf{brown}], [\textbf{tree}, \textbf{larger}], [\textbf{larger}, \textbf{brown}]) where \textbf{larger} and \textbf{brown} represent distractor concepts randomly selected for each task. Workers were required to be located in the U.S., with $>95\%$ HIT acceptance rate and $>100$ HITs accepted. Workers were paid $\$0.06$ per HIT. 

\vspace{-2 pt}
\section{Additional qualitative results}

\begin{figure*}[b!]
\begin{center}
   \includegraphics[width=\linewidth]{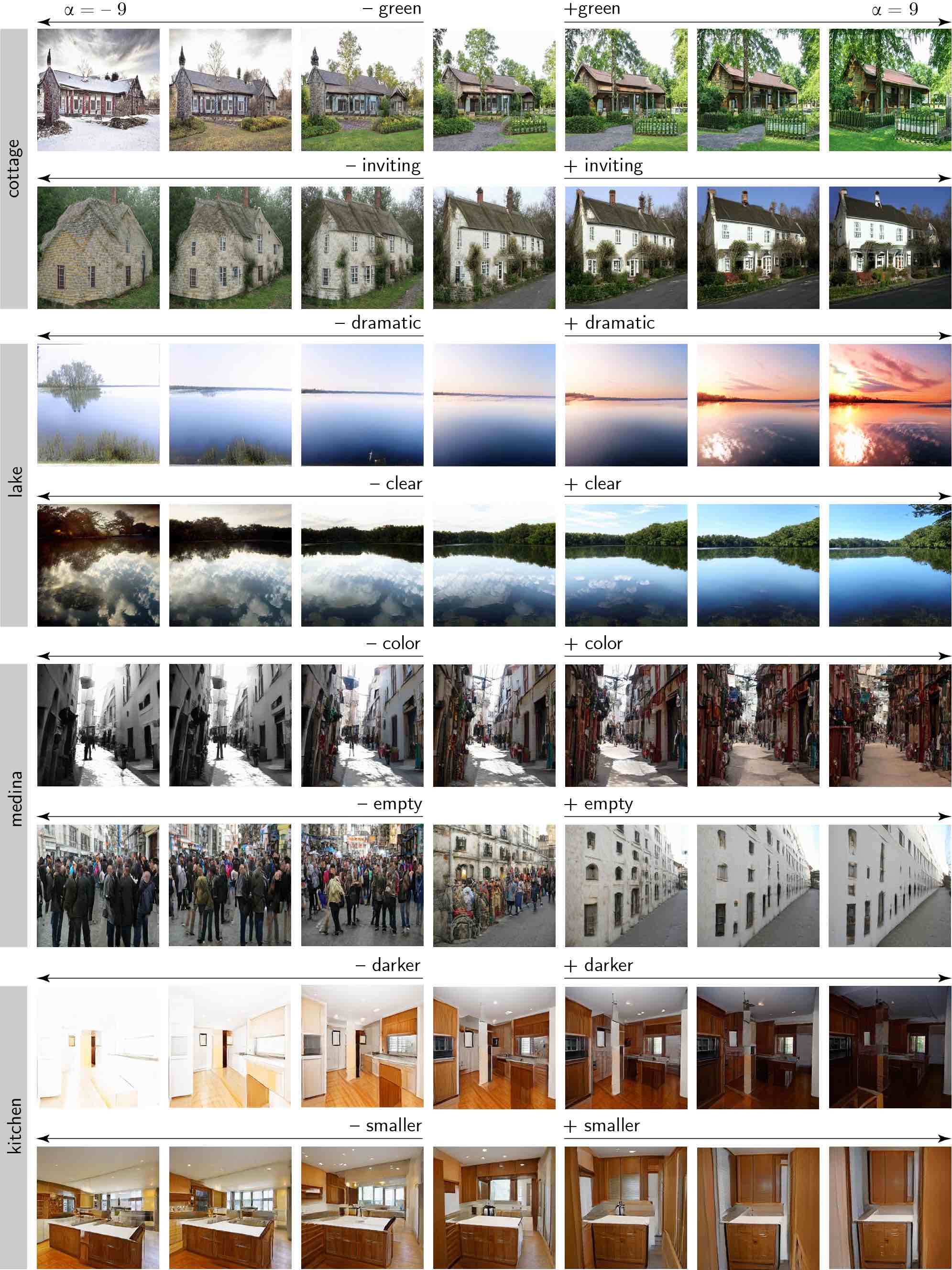}
\end{center}
   \captionsetup{}
   \caption{Additional examples of concept addition and subtraction. $\alpha$ is varied in steps of size 3.}
\label{fig:negatives}
\end{figure*}

\begin{figure*}[b!]
\begin{center}
   \includegraphics[width=\linewidth]{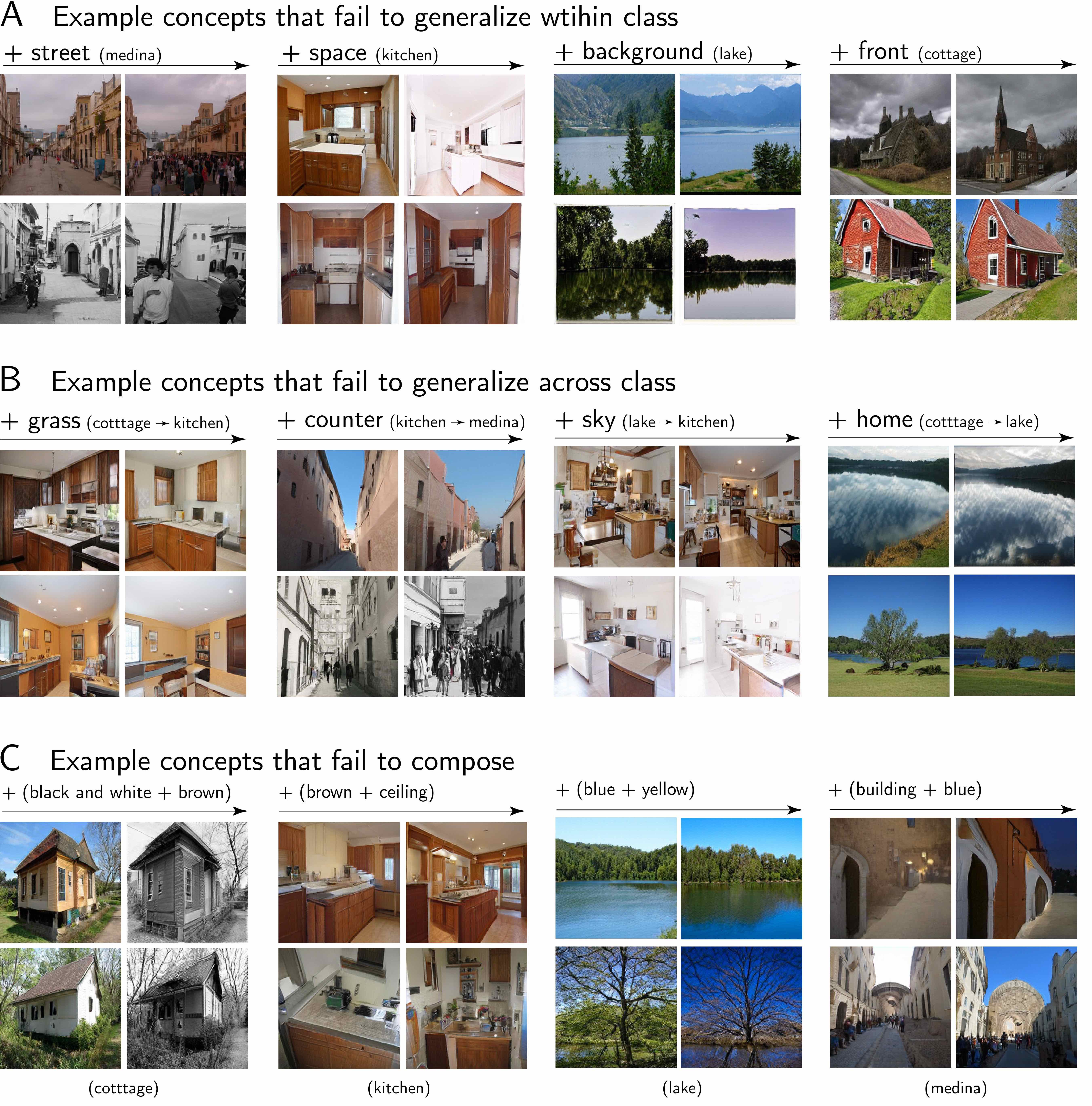}
\end{center}
   \captionsetup{}
   \caption{Example failures. (a) Shows sample concepts that did not perform at accuracy above chance in the AMT task described in Section 4.1. Many of these concepts, such as \emph{space}, are broad in scope and could be used to describe many kinds of scene changes. (b) Shows sample concepts that each performed at accuracy above chance \emph{within class} in the AMT task described in Section 4.2, but failed to perform above chance when tested in a \emph{different class}. (c) Shows concepts that perform above chance individually but not when composed, in the experiment described in Section 4.3.}
\label{fig:failure}
\end{figure*}

 In Figure \ref{fig:negatives} we visualize additional examples of concept subtraction as well as addition for varying $\alpha$ (compare to Figure 4 in the main paper). While most directions generalize across $\mathbf{z}$ and some across $\mathbf{y}$, others do not. Furthermore, Section 4.3 suggests we can compose concepts. This works for concepts that regularly occur in the same class, and some that do not co-occur, but some combinations do not work. In Figure~\ref{fig:failure} we show examples of concepts that fail to generalize across $\mathcal{Z}$ or class, or fail to compose with other concepts. 
\vspace{-20px}